\documentclass[10pt,twocolumn,letterpaper]{article}

\usepackage{mathtools}
\usepackage{cvpr}
\usepackage{times}
\usepackage{epsfig}
\usepackage{graphicx}
\usepackage{amsmath}
\usepackage{amssymb}
\usepackage{hhline}
\usepackage{soul}
\usepackage{amsmath,amssymb} % define this before the line numbering.
\usepackage{color}

\usepackage{makecell}

\usepackage{bm}
\usepackage{array}
\usepackage[normalem]{ulem}
\usepackage{float}
\usepackage{soul}
\usepackage[ruled,vlined]{algorithm2e}
\usepackage{multirow}
\usepackage{subfigure}
\usepackage{ mathrsfs }
\newcolumntype{M}[1]{>{\centering\arraybackslash}m{#1}}
\usepackage{caption}

\usepackage[ruled,vlined]{algorithm2e}

\usepackage[pagebackref=true,breaklinks=true,letterpaper=true,colorlinks,bookmarks=false]{hyperref}

\cvprfinalcopy % *** Uncomment this line for the final submission

\ifcvprfinal\fi
\begin{document}

%%%%%%%%% TITLE
\title{Kinematic-Layout-aware Random Forests for Depth-based Action Recognition}

\author{Seungryul Baek$^1$, Zhiyuan Shi$^1$, Masato Kawade$^2$, Tae-Kyun Kim$^1$\\
$^1$ Imperial College London, $^2$ OMRON Corporation \\
{\tt\small \{s.baek15,z.shi\}@imperial.ac.uk, kawade@ari.ncl.omron.co.jp, tk.kim@imperial.ac.uk}
%\and
}

\maketitle
%\thispagestyle{empty}

%%%%%%%%% ABSTRACT
\begin{abstract}
%Existing works for action recognition from depth sequences focus on modeling the dynamics of various actions. 
In this paper, we tackle the problem of 24 hours-monitoring patient actions in a ward such as 
%``lying on the bed'', 
``stretching an arm out of the bed'', ``falling out of the bed'', where temporal movements are subtle or significant. In the concerned scenarios, the relations between scene layouts and body kinematics (skeletons) become important cues to recognize actions; however they are hard to be secured at a testing stage. 
%thus their relative positions to scene layouts and body postures become important cues. 
%{; however both are not secured at testing due to several challenges (\eg difficult human poses, viewpoint variation, cluttered scenes)}. 
To address this problem, we propose a kinematic-layout-aware random forest which takes into account the kinematic-layout (\ie layout and skeletons), to maximize the discriminative power of depth image appearance.
%{Both appearance and kinematic information are encoded considering view-invariance, which is one of important issues in a depth problem.} 
We integrate the kinematic-layout in the split criteria of random forests to guide the learning process
%, \ie. the spatio-temporal appearance feature selection, 
by 1) determining the switch to either the depth appearance or the kinematic-layout information, and 2) implicitly closing the gap between two distributions obtained by the kinematic-layout and the appearance, when the kinematic-layout appears useful.
%4) clustering synthetic multi-view data randomly. 
The kinematic-layout information is not required for the test data, thus called ``privileged information prior''.
%and 
%{the information based on the ground-truth is used at the training only}, 
%thus called ``privileged information prior''.}
%The learned relationship from the kinematic space to the appearance space helps maximize class separation of \st{both static and dynamic} actions. 
The proposed method has also been testified in cross-view settings, by the use of view-invariant features and enforcing the consistency among synthetic-view data. 
 %across \seungryul{responses.} 
Experimental evaluations on our new dataset PATIENT, CAD-60 and UWA3D (multiview) demonstrate that our method outperforms various state-of-the-arts.
\end{abstract}

\vspace{-0.5cm}
%%%%%%%%% BODY TEXT
\section{Introduction}

The recent emergence of cost-effective and easy-operation depth sensors have opened the door to a new family of methods \cite{Omar_cvpr_2013,Lu_cvpr_2013,Xiaodong_2014_cvpr,Jiajia_iccv_2013,RFaction,Cewu_cvpr_2014,CVPR15_heterogeneous,Kong_2015_cvpr,hanklett} for action recognition from depth sequences. Compared to conventional color images, depth maps offer several advantages: 1) Depth maps encode rich 3D structural information, including informative shape, boundary, geometric cues of a human body and an entire scene. 2) Depth maps are insensitive to changes in lighting and illumination conditions that make it possible to monitor patient/animal 24/7. 3) It is invariant to texture and color variations, which eases the task of human detection and segmentation.
\begin{figure}[t]
\centering
   \includegraphics[width =\linewidth] {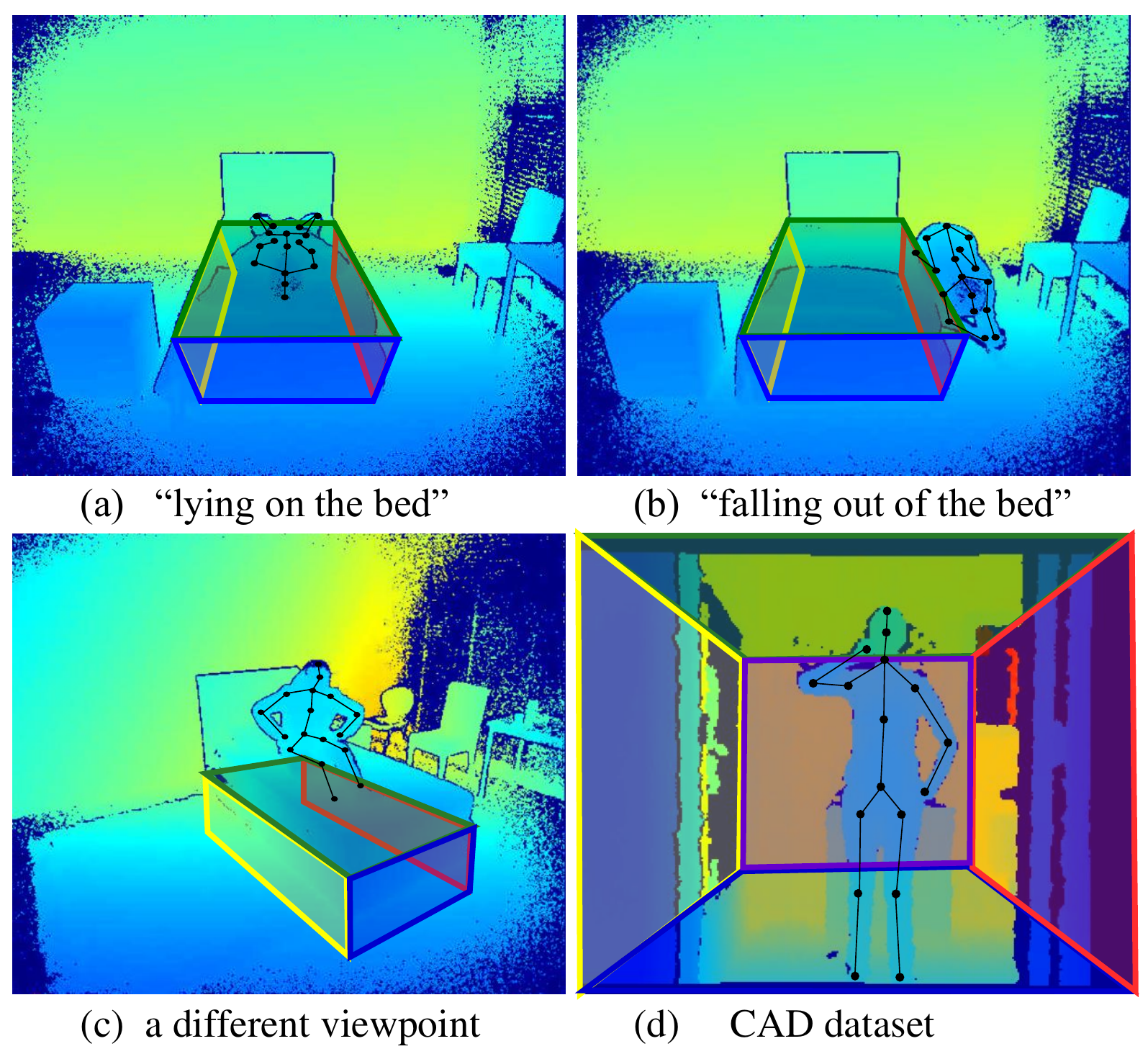}
 
\caption{Examples of datasets with kinematic-layout information. (a) ``lying on the bed'' action, (b) ``falling out of the bed'' action, (c) a different viewpoint, in our dataset, and (d) CAD dataset.}
\label{fig:our_data}
\end{figure}

These advantages have promoted the fast pace development of depth-based techniques for action recognition. A number of spatio-temporal representations \cite{Laptev_2005_ijcv,Omar_cvpr_2013} have been proposed to handle the challenges of depth discontinuities and noise. When human skeleton can be estimated from depth sequences, recent approaches resorted to selecting the informative points around skeleton joints and modelling their temporal dynamics~\cite{Jiang_tpami_2014,Wei_2013_ICCV}. 
The above methods are not well suited to either static actions or difficult body poses with severe occlusions as in Fig. \ref{fig:our_data}(a) and \ref{fig:our_data}(b). Solving the task of action recognition and pose estimation jointly \cite{Yu_2013_cvpr,Yao_2011,Nie_2015_CVPR,Gall2010} has attracted attention as they are closely related tasks for understanding human motion. They perform both pose and action recognition at testing, in which accurate action recognition is conditioned on reasonable pose estimation. More recently, Fouhey \etal \cite{Fouhey2014} and Delaitre \etal \cite{Fouhey_2012} show that the coupling between human actions and scene geometry provides a strong cue for scene understanding. Inspired by this observation, we aim to maximize action discrimination by exploiting kinematic-layout information prior (\eg room layout, body skeletal parts).
%\seungryul{However, human skeletons are not well secured when humans are not in the upright positions \cite{HOPC1,HOPC2}. Due to novel viewpoints and difficult human poses}
Considering that this information prior, in particular the skeletal parts, are difficult to obtain in our scenarios (\eg when the person lies on the bed), and when dealing with test videos from unseen camera viewpoints as in Fig \ref{fig:our_data}(c). 
%\seungryul{Also, the layout estimation suffers in cluttered indoor scenes \cite{Layoutdiff,Hedau_iccv_2009,ECCV2010_layout}.} 
We seek to formulate this information as privileged knowledge \cite{Yang_tcsvt_2015,Vapnik2009544} that is only required during training. At the testing stage, we directly apply our method to raw depth sequences.

\begin{figure}[t]
\centering
   \includegraphics[width =\linewidth] {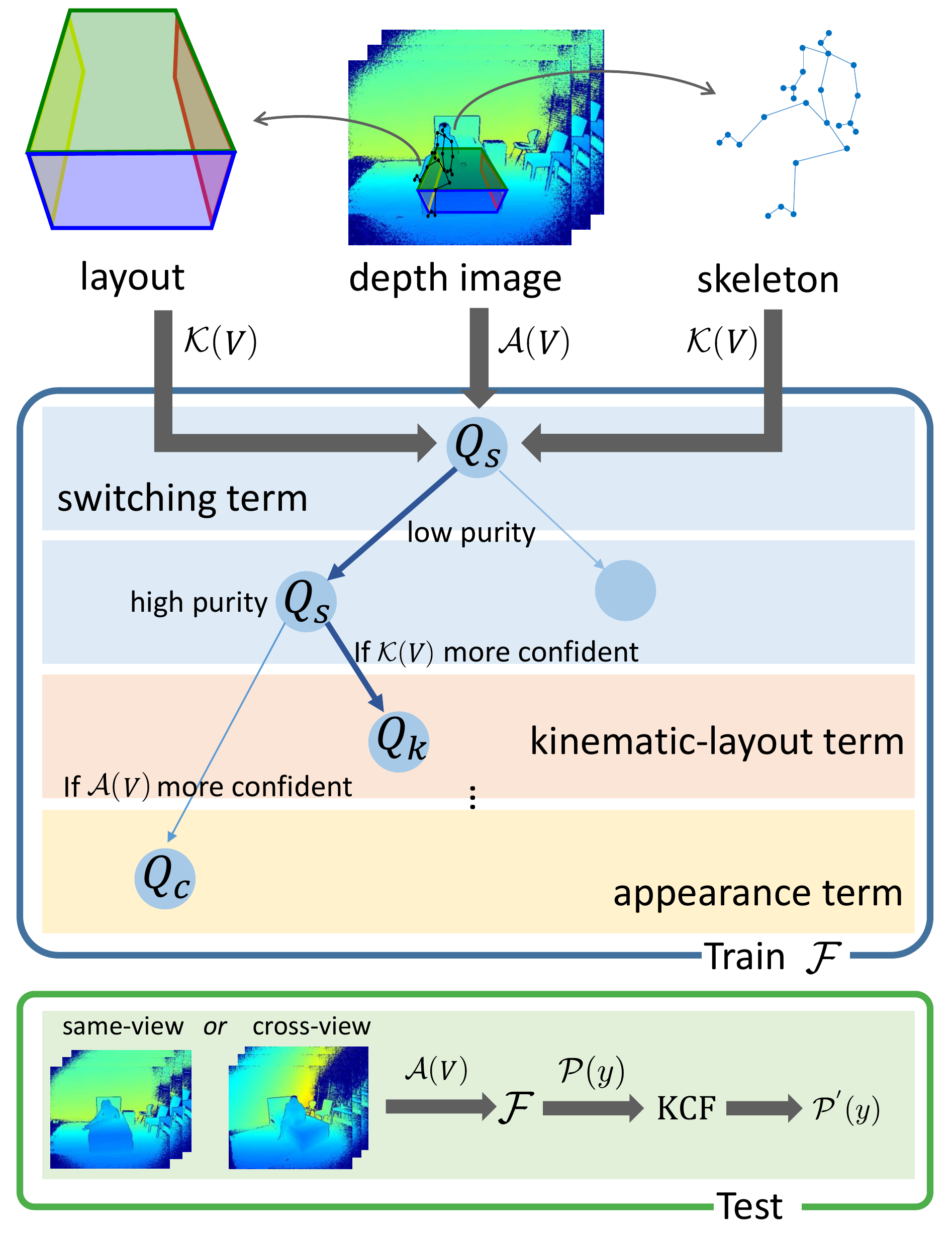}
 
\caption{\textbf{Flowchart of our method for action recognition.} The model $\mathcal{F}$ takes as an input, depth appearance $\mathcal{A}(V)$ and kinematic-layout information $\mathcal{K}(V)$ in training. Quality functions ($Q_s$,$Q_c$,$Q_k$) are designed to optimize the model $\mathcal{F}$, by considering both depth appearance $\mathcal{A}(V)$ and kinematic-layout information $\mathcal{K}(V)$ (refer to the text for detailed explanations). The trained model $\mathcal{F}$ is directly applied to raw-depth sequences $\mathcal{A}(V)$ at testing, in both single-view and cross-view experiments. Kinematic consistency filter (KCF) produces the final class distribution $P'(y)$, by refining class distribution $P(y)$ from $\mathcal{F}$.}
\label{fig:our_model}
\end{figure}

In order to investigate these issues, in this paper we make the following contributions: 

\vspace{0.2cm}
\noindent \textbf{New action recognition dataset (PATIENT)} has been collected in the scenario of 24 hours-monitoring patient behaviors (15 actions) in a ward by a depth camera. Compared to the conventional action recognition tasks \cite{Lu_cvpr_2013}, actions of patients 
%\seungryul{are in novel viewpoint where human pose estimation is hard} 
contain both subtle (\eg ``lying on the bed'', ``stretching an arm out of the bed'') and notable temporal movements (\eg ``falling out of the bed'') (see Fig. \ref{fig:our_data} and \ref{fig:our_dataset}), where kinematic-layout information is helpful, under challenging viewpoints. Unlike existing datasets, skeleton information of our data can not be reliably tracked by a depth sensor (\eg kinect) due to a special viewpoint and severe occlusions.

\vspace{0.1cm}
\noindent \textbf{Information prior} is defined to help a model incorporate both the kinematic-layout information and the depth appearance. The kinematic-layout is treated as the privileged information in the training stage. 
%\seungryul{Both appearance and kinematic features are encoded considering view-invariance.} 
The privileged information mainly comes from two sources: (1) the scene layout of the monitored environment; (2) human skeleton or key points. The scene layout cue reveals a geometric relationship between scene layouts and human actions. For example, ``lying on the bed'' implies the overlap area between a human actor and bed; ``writing on a whiteboard'' indicates the relative position of a human actor. The skeleton cue provides the location and kinematics of human body. The privileged information is exploited to preserve the coherence among geometry, kinematic structure and depth appearance, which aims to enhance the discriminative power of appearance cue for action classification. 
%The learned model can better explain the depth evidence to correctly recognize the action, especially for those more static actions. However, this additional information and its estimation are not required for test data. (except the smoothing in Sec. 3.3)

\vspace{0.1cm}
\noindent \textbf{Kinematic-layout-aware random forest (KLRFs)} is introduced to improve the discriminative power of the depth appearance by selectively encoding the kinematic-layout information. The switching term $Q_s$ first clusters data samples into two groups: kinematic-layout based and appearance-based samples. Then, kinematic-layout term $Q_k$ and appearance term $Q_c$ are applied adaptively to each group to split them effectively. Kinematic-layout information is implicitly used at training to guide the learning process and not required at testing (See Fig. \ref{fig:our_model}).

%This information is formulated in two ways to guide the learning process both explicitly and implicitly: (1) Our method directly measure the uncertainty of kinematic information by embedding the statistics of layout-skeleton and skeleton-skeleton to a split criterion. This term learns the regression aspect of the random forest to improve feature selection for the class separation. (2) This kinematic information can also be considered as input feature vectors to train a separate prior forest. Data samples (\ie. depth images) are weighted to implicitly close the gap between the posterior class distributions of the kinematic cue and depth appearance features in an active selection manner. 

\vspace{0.1cm}
\noindent \textbf{Both cross and single-view settings} are experimented to demonstrate the generalization of the proposed method. We evaluate our approach on the proposed PATIENT dataset, Cornell Activity Dataset, and UWA3D Multiview Activity II dataset. Kinematic consistency filter (KCF) is applied to aggregate responses of augmented synthetic view data and to infer the final result. View clustering term is further proposed to deal with cross-views. Also, depth appearance is encoded in a view-invariant fashion~\cite{novelview}. The extensive experiments demonstrate that our approach provides more accurate results compared to the state-of-the-art methods.  

%d object recogni- tion across different viewpoints, are conducted to evaluate the usefulness and generalizability of the approach. Supe-

%We encode both kinematic information and appearance features in a view-invariant representation to . 

%\seungryul{Our appearance feature is view-invariant due to finetuning with rotated/translated synthetic data and Kinematic-layout-aware view clustering term of our forests. Our kinematic feature becomes invariant to translational and view variations due to plane distance and min/max encoding method.}

\setlength{\belowcaptionskip}{-16pt}

\section{Related works}
\label{sec:relate}

In this section, we review prior works on depth-based action recognition and random forest with information prior. We discuss the difference between our model and relevant techniques.

\subsection{Depth-based action recognition}

Color and texture are precluded in depth sequences, which enlightens the existing work to explore the following information cues:
  
\noindent \textbf{Spatio-temporal cue.} \ Spatial cue captures the static appearance information of single frames. Temporal cue conveys the movement of the observee or objects in the form of motion across frames. These two cues are usually encoded together as a spatio-temporal representation. The interest point detection and description has been widely studied \cite{Vemuri_1986,Pauly_2003,Rusu_2010} to provide reliable features for describing humans, objects or scenes. The spatio-temporal interest points (STIPs) are often adopted \cite{Laptev_2005_ijcv,Dollar_2005,Willems_2008_eccv,Laptev_2008_CVPR,klaser_bmvc_2008} for compact representations of activities and events. These conventional RGB-based methods do not perform well on depth maps \cite{Dollar_2005,Heng_2011_cvpr,Shandong_iccv_2011,Laptev_2005_ijcv}. Recent efforts \cite{Jiang_eccv_2012,Vieira_2012,Lu_cvpr_2013,Zhang_2011,Wanqing_cvprw_2010,Omar_cvpr_2013,Xiaodong_2014_cvpr}, therefore, have been devoted to developing reliable interest points and tracks for depth sequences. The interest points are extracted from low-level pixels \cite{Wanqing_cvprw_2010,Bingbing_2011_iccv,Cheng_eccv_2012} or mid-level parts \cite{Yang_cvpr_2015,Cewu_cvpr_2014,Sadanand_cvpr_2012}. In contrast to using local points, a holistic representation \cite{Yang_2012_acm,Vieira_2012,Wanqing_cvprw_2010,Jiang_eccv_2012} is recently popular as it is shown generally effective and computationally efficient. Yang \etal \cite{Yang_2012_acm} extracted Histograms of Oriented Gradients (HOG) descriptors from Depth Motion Maps (DMM), where the DMM are generated by stacking motion energy of depth maps projected onto three orthogonal Cartesian planes. Wang \etal \cite{wang2015cnn} defined Hierarchical Dynamic Motion Maps (HDMM) by using different offsets between frames and extracting Convolutional Neural Network (CNN) features from them. More recently, Rahmani \etal proposed a view-invariant descriptor HOPC \cite{HOPC2} to deal with the 3D action recognition from unseen views \cite{novelview}.

%HOPC descriptor \cite{HOPC2} and CNN-based view-invariant feature extractor \cite{novelview} to relieve it.} \seungryul{View-invariance is an issue for depth-based cues. 

\noindent \textbf{Skeleton/pose cue.} Pose estimation is beneficial for understanding human actions \cite{Yao_2011,Gall2010,hanklett}, while action recognition can also facilitate 3D human pose estimation \cite{Yu_2013_cvpr}. The joint modeling of action and pose has been studied on RGB data \cite{Nie_2015_CVPR,Ukita_2013,Lv_cvpr_2007,cheronICCV15,PSSR1,PSSR2}. They perform pose estimation at testing stages, which either helps further action recognition or is helped by prior action recognition. In either case, accurate pose estimation at testing is aimed. A well trained skeleton tracker can provide a high-level cue for depth sequences. The use of skeleton joints has been suggested by \cite{Jiang_tpami_2014,Xia_cvor_2012} for alleviating ambiguities in action recognition. Jiang \etal \cite{Jiang_tpami_2014} represent the interaction between human body parts and environmental objects with an ensemble of human joint-based features. Skeleton joints have also been used to constrain the dictionary learning for feature representation \cite{Jiajia_iccv_2013}. Although they were shown to yield high recognition accuracies, the estimated 3D joint positions are not always stable due to the noisy depth maps \cite{Zanfir_2013_ICCV,Chunyu_cvpr_2013}. Zanfir \etal \cite{Zanfir_2013_ICCV} present a representation that captures not only 3D body poses but also differential properties (\ie. speed, acceleration). Wang \etal \cite{Chunyu_cvpr_2013} consider the best-$K$ joint configurations to reduce the joint estimation errors.  %They show the improved accuracy, even RGB and skeletons are obtained from different label spaces \ie skeletons are used in either unsupervised or semi-supervised manner. }

\noindent \textbf{Layout cue.} Scene layout provides a geometry information about visible surfaces of object, wall, floor, and ceiling \cite{Hedau_iccv_2009,ECCV2010_layout}. Fouhey \etal \cite{Fouhey2014} and Delaitre \etal \cite{Fouhey_2012} show that by observing human behavior, a strong correlation can be found between human actions and properties of a scene and its objects. Similarly, Savva \etal \cite{Savva_2014_tog} observe and track people as they interact with the environment using RGB-D sensors. These methods aim at improving the estimation of 3D scene geometry.

Most previous studies focus on representing and modeling the temporal dynamics of human actions. While some works \cite{Gkioxari_2015_cvpr,NIPS2014_twostream,couprie_iclr_13} have attempted to learn spatial cues to capture the static appearance of color images, it is still not straightforward to generalize them for depth images. In this work, we propose a new dataset to explore representations and learning schemes for recognizing more static as well as dynamic human actions from depth sequences. Recently, Mahasseni \etal \cite{reglstm} proposed to exploit skeleton information as a regularizer, when training a long short term memory architectures for RGB-based action recognition. We aim to capture both static and dynamic action appearance by exploiting both kinematic and layout cues selectively according to its effectiveness. Due to the challenging and unseen viewpoints, kinematic-layout information is encoded as privileged information \cite{Yang_tcsvt_2015,Vapnik2009544}, which is only required during training. We do not require the estimation of such kinematic-layout information at the testing stage.

%\begin{figure}[t]
%\centering
   %\includegraphics[scale =0.3] {images/layout_plane.pdf}
%    \includegraphics[width=\linewidth] {layout_plane.pdf}
%\caption{Examples of labeled layout planes $\mathbb{L}$ in (a) PATIENT (b) CAD60 dataset.}
%\label{fig:layout_plane}
%\end{figure}

\subsection{Random forest with information prior}

Standard random forests make the assumption that the output variables are independent over the parameter space. Conditional regression forest was presented by Sun \etal \cite{Sun_cvpr_2012} and Dantone \etal \cite{dantone_2012_cvpr}, which demonstrates that the incorporation of prior information (such as human height, head pose) can enhance the dependency between output variables and latent variables, resulting in more accurate predictions. Similarly, Dapogny \etal \cite{Dapogny_2015_ICCV} and Pham \etal \cite{Pham_2015_ICCV} utilize expression prior and crowdedness prior respectively to reduce the variability within classes. Our method differs from existing conditional forests in that most of them exploited prior to model the probability functions over the leaf nodes, we utilize the prior information at the split nodes while growing trees. 

Learning with privileged information (LUPI) \cite{Vapnik2009544} shares a similar spirit in utilizing additional information at the training stage. This information is only available during training, which provides better explanations of the data. Tang \etal \cite{hrf_iccv2013} use additional synthetic data to establish associations with real data. Yang \etal \cite{Yang_tcsvt_2015} exploited the discrete additional prior explicitly to improve the quality of decision trees. Our method incorporates continuous prior to guide the learning process both explicitly and implicitly.

\section{Kinematic-layout-aware random forest}
\label{sec:our_model}
%\subsection{Method overview}
Random forests (RFs) $\mathcal{F}$ aim to learn a nonlinear mapping from the appearance $\mathcal{A}$ to the label set $\mathcal{Y}$ :
\begin{eqnarray}
\label{RFmapping}
\mathcal{F}: \mathcal{A} \mapsto \mathcal{Y}. 
\end{eqnarray}
RFs $\mathcal{F}$ are ensembles of binary trees, containing two types of nodes: {\it split} and {\it leaf}. Split nodes decide its input $V$ goes either to the left child (if $\mathcal{A}(V)^{\gamma}<\tau$) or to the right child (otherwise) according to its split function $\Psi(\mathcal{A}(\cdot)^{\gamma}, \tau)$ where $\mathcal{A}(\cdot)^{\gamma}$ denotes the $\gamma$-th value in the appearance feature and $\tau$ is a threshold. Leaf nodes are terminating nodes, which store statistics of training samples (\eg class label $y\in\mathcal{Y}$ or regression vectors). At training stage, trees are grown by deciding the split function $\Psi(\mathcal{A}(\cdot)^{\gamma}, \tau)$ recursively from the root node. At each node, the arrived input data $V\in \mathcal{D}$ is divided into two subsets $\mathcal{D}_l$ and $\mathcal{D}_r$ ($\mathcal{D}_l \cap \mathcal{D}_r = \emptyset $) by a set of split function candidates $\{\Psi^c\}$ that is generated randomly. Among candidates, the one that maximizes the quality function $\mathcal{Q}$ is selected as a split function $\Psi^*$ as follows:
\begin{eqnarray}
\label{optimizeQ}
\Psi^*=\arg\max_{\Psi\in\{\Psi^c\}} \mathcal{Q}(\Psi)
\end{eqnarray}
where the quality function $\mathcal{Q}$ is defined based on tasks (\eg entropy for classification, variance for regression). Trees are grown until only $1$ sample is remained in the node or no information gain is obtained, where the information gain is defined as $\mathcal{Q}(\Psi^*)-\mathcal{Q}(\Psi^0)$ where $\Psi^0$ is the reference split that have all samples in $\mathcal{D}_l$ and no samples in $\mathcal{D}_r$.

\noindent \textbf{Overview of the proposed method.} We propose kinematic-layout-aware random forests (KLRFs) $\mathcal{F}^+$ to optimize the nonlinear mapping in Eq. \ref{RFmapping} with the help of the kinematic-layout $\mathcal{K}$ in training, when it is useful : 
\begin{eqnarray}
\mathcal{F}^{+}: \left\{ \begin{array}{ll}
\mathcal{A} \xmapsto{\mathcal{K}} \mathcal{Y}, & \textrm{, if $\mathcal{K}$ is useful}\\
\mathcal{A} \mapsto \mathcal{Y} & \textrm{, otherwise}
\end{array} \right.
\label{KLRFmapping}
\end{eqnarray}
where our appearance $\mathcal{A}$ is defined as depth maps and the kinematic-layout $\mathcal{K}$ is defined as scene layouts, skeleton joints and their associations (see Sec.~\ref{sec:priors}). For the purpose, we propose to use different quality functions $\mathcal{Q}=\{Q_s,Q_c,Q_k\}$ (see Sec.~\ref{sec:forest}) by combining different information sources : 
\begin{eqnarray}
\label{qs}
\mathcal{Q}=\left\{\begin{matrix}
Q_s(\Psi;\mathcal{A},\mathcal{K},\mathcal{Y}): \textrm{switching term}
\\  Q_c(\Psi;\mathcal{A},\mathcal{Y}): \textrm{appearance term}
\\ Q_k(\Psi;\mathcal{A},\mathcal{K},\mathcal{Y}): \textrm{kinematic-layout term}
%\\ \mathcal{Q}_v(\Psi;\mathcal{A},\mathcal{K})
\end{matrix}\right.
\end{eqnarray}
Thanks to the random and hierarchical nature of RFs, we are able to utilize different quality functions within a forest, by switching among them at split nodes. Note that split functions $\{\Psi(\mathcal{A}^{\gamma}(\cdot),\tau)\}$, same as standard RFs, operate based only on appearance $\mathcal{A}$, and thus, once RF training is done, the kinematic-layout $\mathcal{K}$ is not used at testing stages.

In following subsections, we first introduce our appearance and kinematic-layout information (Sec.~\ref{sec:priors}) and then present how our approach adaptively selects quality functions to exploit both kinematic-layout and appearance information (Sec.~\ref{sec:forest}). Testing stage of KLRFs and cross-view setting are explained in Sec.~\ref{sec:test} and Sec.~\ref{sec:cross}, respectively.

%RF method that exploits both kinematic-layout information and depth appearance (Sec.~\ref{sec:forest}).

\subsection{Appearance and kinematic-layout information}
\label{sec:priors}
%\vspace{-0.1cm}
We define the appearance $\mathcal{A}$ based on the raw-depth sequences, while define the kinematic-layout $\mathcal{K}$ based on scene layout and skeleton cues. Since it is hard to secure scene layouts and skeleton cues automatically in our scenarios, we exploit their manual ground-truths only for training stages. 1) We first extract the depth cue $\mathbf{C}^D_t$, scene layout cue $\mathbf{C}^L_t$ and skeleton cue $\mathbf{C}^J_t$ for each frame $t$. 2) Then, we generate the spatio-temporal representation, $\mathcal{A}(V)$ and $\mathcal{K}(V)$ for a depth sequence $V$, by applying the Fourier transform on per-frame cues as in \cite{novelview,Jiang_tpami_2014}. Individual frame representations are defined as follows:\\
\noindent \textbf{Depth cue $\mathbf{C}^D_t$:} For each frame $t$, we extract the $4,096$ dimensional feature $\mathbf{C}^D_t$ by a pre-trained CNN architecture~\cite{novelview} on depth images. This architecture is pre-trained on synthetic multi-view depth maps and shown to produce the state-of-the-art accuracy on both single and multi-viewed 3D action recognition benchmarks~\cite{novelview}.

\noindent \textbf{Scene layout cue $\mathbf{C}^L_t$:} There exists a strong physical and functional coupling between human actions/poses and the 3D geometry of a scene \cite{Fouhey2014,Fouhey_2012}. For each frame $t$, we extract a descriptor $\mathbf{C}^L_t$ by 3D displacements between scene layouts $\mathbb{L}=\{\mathbf{L}_1,...\mathbf{L}_l,...,\mathbf{L}_L\}$ and 3D human skeleton joints $\mathbb{P}(t)=\{\mathbf{p}_1(t),...\mathbf{p}_p(t),...,\mathbf{p}_P(t)\}$ as:
%\vspace{-0.1cm}
\begin{eqnarray}
\label{skleq}
\mathbf{C}^{L}_t=[\mathbf{d}_{t11};...;\mathbf{d}_{t1L};\mathbf{d}_{t21};...;\mathbf{d}_{t2L};...;\mathbf{d}_{tP1};...;\mathbf{d}_{tPL}]
\end{eqnarray}
where %\begin{eqnarray}
$\mathbf{d}_{tpl} = \mathbf{p}_p(t)-\mathbf{\bar{p}}_{\mathbf{L}_l}$
%\end{eqnarray}
, $\mathbf{p}_p(t)$ is a 3-dimensional vector whose entry corresponds to its x, y and depth value and $\mathbf{\bar{p}}_{\mathbf{L}_l}$ is a projection of $\mathbf{p}_p(t)$ to the plane $\mathbf{L}_l$, respectively. Due to the plane-to-point distance, distances between skeleton joints and layout planes are translation invariant.
%To further deal with the view variation, we divide entire layouts $\mathbb{L}$ into two groups: $\mathbb{L}_{side}$ and $\mathbb{L}_{normal}$, where the former is a layout group whose entries are considered together for view-invariance while the latter is a layout group whose entries are considered individually. The new representation is defined as follows:
%\begin{eqnarray}
%\label{skleq}
%\mathbf{C}^{L}_t=[\mathbf{C}^{L'}_t;\mathbf{d}_{t1}^{\max};\mathbf{d}_{t1}^{\min};...;\mathbf{d}_{tP}^{\max};\mathbf{d}_{tP}^{\min}],
%\end{eqnarray}
%where $L'$ is the number of elements in $\mathbb{L}_{normal}$ and
%\begin{eqnarray}
%\mathbf{d}_{tp}^{\max} &= |\mathbf{p}_p(t)-\mathbf{\bar{p}}_{\mathbf{L}_{l^{\max}}}|,\\
%\mathbf{d}_{tp}^{\min} &= |\mathbf{p}_p(t)-\mathbf{\bar{p}}_{\mathbf{L}_{l^{\min}}}|
%\end{eqnarray}
%where
%\begin{eqnarray}
%l^{\max} &= \arg\max_{l\in\mathbb{L}_{side}}||\mathbf{p}_p(t)-\mathbf{\bar{p}}_{\mathbf{L}_l}||,\\
%l^{\min} &= \arg\min_{l\in\mathbb{L}_{side}}||\mathbf{p}_p(t)-\mathbf{\bar{p}}_{\mathbf{L}_l}||.
%\end{eqnarray}
%As a result, for $\mathbb{L}_{normal}$, $\mathbf{d}_{tpl}$ in Eq. \ref{skleq} is obtained for each individual layout. For $\mathbb{L}_{side}$, only maximum and minimum distances $\mathbf{d}_{tp}^{\max}$, $\mathbf{d}_{tp}^{\min}$ are obtained for all entries in $\mathbb{L}_{side}$.
%Note that layout planes $\mathbb{L}$ are the landmarks whose locations do not change across the temporal index $t$.
This layout cue provides information on how humans interact with their environments. Some actions, such as ``sitting'' and ``lying'', are supported by certain planes considering physical constraints. %This layout information can be defined in different ways. (\eg one might measure perpendicular distances of skeletal parts to the planes).

\noindent \textbf{Skeleton cue $\mathbf{C}^J_t$:} Skeleton cue $\mathbf{C}^J_t$ is further encoded in three ways: $\mathbf{C}^J_t=[\mathbf{d}^{P}_t; \mathbf{d}^{M}_t; \mathbf{d}^{O}_t]$. (1) Skeleton {\it Pairwise} distance vector, $\mathbf{d}^{P}_t=[\mathbf{p}_1(t)-\mathbf{p}_2(t),...,\mathbf{p}_p(t)-\mathbf{p}_q(t),...,\mathbf{p}_{P-1}(t)-\mathbf{p}_P(t)]$ is defined for $\forall p, \forall q, p\ne q \in [1,P]$ to encode current frame's human poses. (2) Skeleton {\it Motion} vector \cite{CVPR2013_daction2}, $\mathbf{d}^M_t=[\mathbf{p}_1(t)-\mathbf{p}_1(t-1),...,\mathbf{p}_p(t)-\mathbf{p}_p(t-1),...,\mathbf{p}_P(t)-\mathbf{p}_P(t-1)]$ is defined for $\forall p\in[1,P]$ to encode its temporal motion information. (3) Skeleton {\it Offset} vector, $\mathbf{d}^{O}_t=[\mathbf{p}_1(t)-\mathbf{p}_1(1),...,\mathbf{p}_p(t)-\mathbf{p}_p(1),...,\mathbf{p}_P(t)-\mathbf{p}_P(1)]$ is defined for $\forall p\in [1,P]$ to encode human offset information to their initial values \ie $t=1$. Skeleton cue can consider the spatial location of human body parts. %\st{Note that the skeleton cue is less informative in PATIENT dataset, since full-body skeletons are not obtained (as in CAD60 dataset).}

\subsection{Learning kinematic-layout-aware forests}
\label{sec:forest}
%Kinematic vectors $\mathcal{K}(V)$ can also be stored to use it for enforcing consistency across frames (see Sec.~\ref{sec:pp}).
\begin{figure}[t!]
\centering
%\subfigure[Example of grown trees]{
%   \includegraphics[width=0.95\linewidth] {forest.pdf}
%   \label{fig:subfig1split}   
% }
%\subfigure[Weighting Method]
%{
   \includegraphics[width=1.1\linewidth] {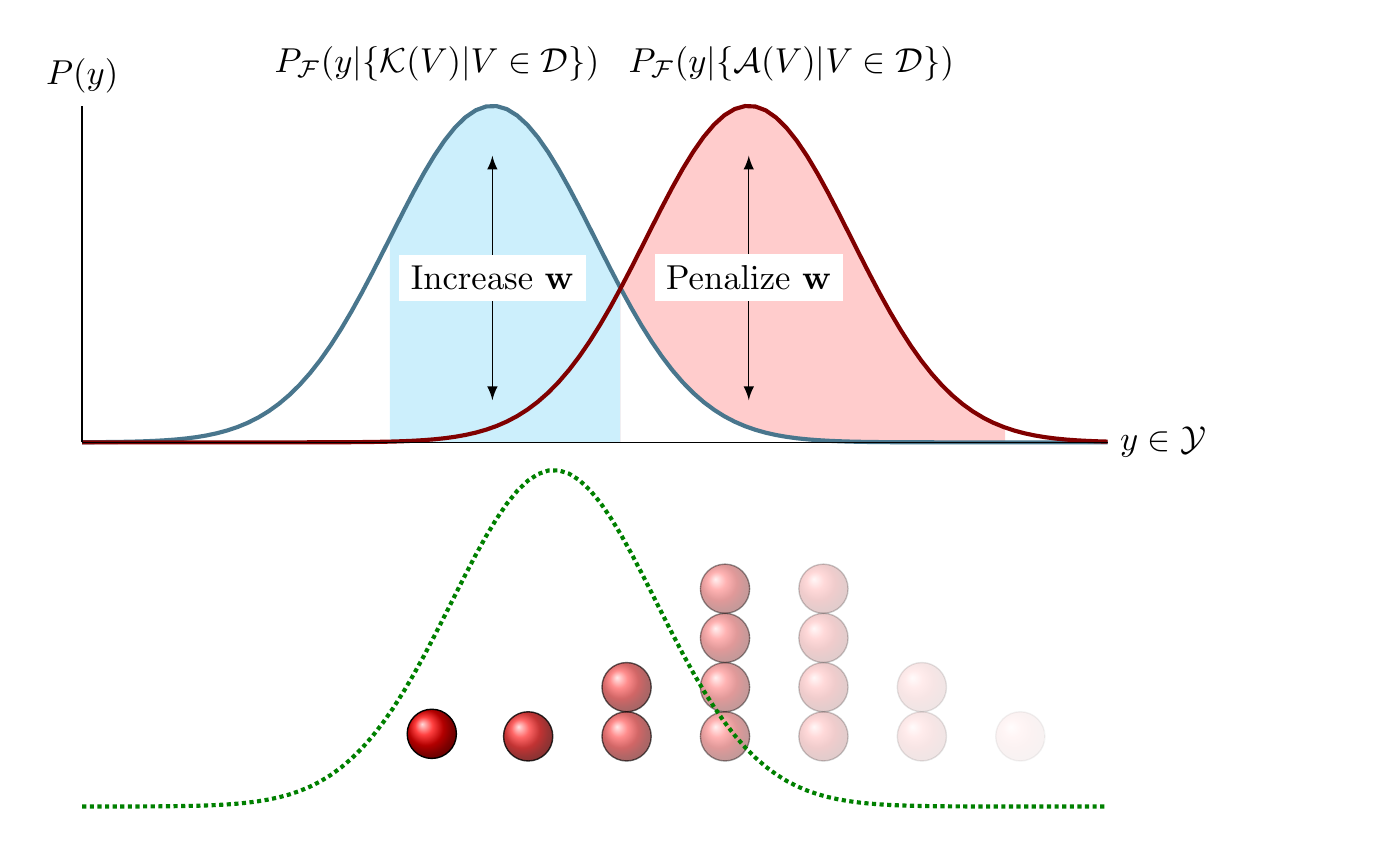}
   %\label{fig:subfig2split}   
%}
\caption{The weighting method to reduce the gap between $P_{\mathcal{F}}(y|\{\mathcal{A}(V)|V\in\mathcal{D}\})$ and $P_{\mathcal{F}}(y|\{\mathcal{K}(V)|V\in\mathcal{D}\})$. Red balls denote samples constituting the appearance-based distribution $P_{\mathcal{F}}(y|\{\mathcal{A}(V)|V\in\mathcal{D}\})$ with their weights in fade-out. Green line denotes the gap-reduced class distribution.}
\label{fig:treegrow}
\end{figure}

To train KLRFs $\mathcal{F}^+$ as in Eq. \ref{KLRFmapping}, we propose to use different types of quality functions as in Eq. \ref{qs}. They are combined into a quality function $\mathcal{Q}$ by variables $\alpha, \beta$ as follows:
\begin{eqnarray}
%\mathcal{Q}(\Psi) = \left\{ \begin{array}{ll}
%\alpha Q_s+(1-\alpha) \{ \beta Q_c+(1-\beta) Q_k\} & \textrm{, if $\gamma>0.5$}\\
\mathcal{Q}(\Psi)=\alpha Q_s+(1-\alpha) \{ \beta Q_c+(1-\beta) Q_k\}
% Q_v & \textrm{, otherwise}
%\end{array} \right.
\end{eqnarray}
where $Q_s$, $Q_c$ and $Q_k$ are switching term, appearance term and kinematic-layout term, respectively. 

\begin{figure*}[t]
\centering
   \includegraphics[width =\linewidth] {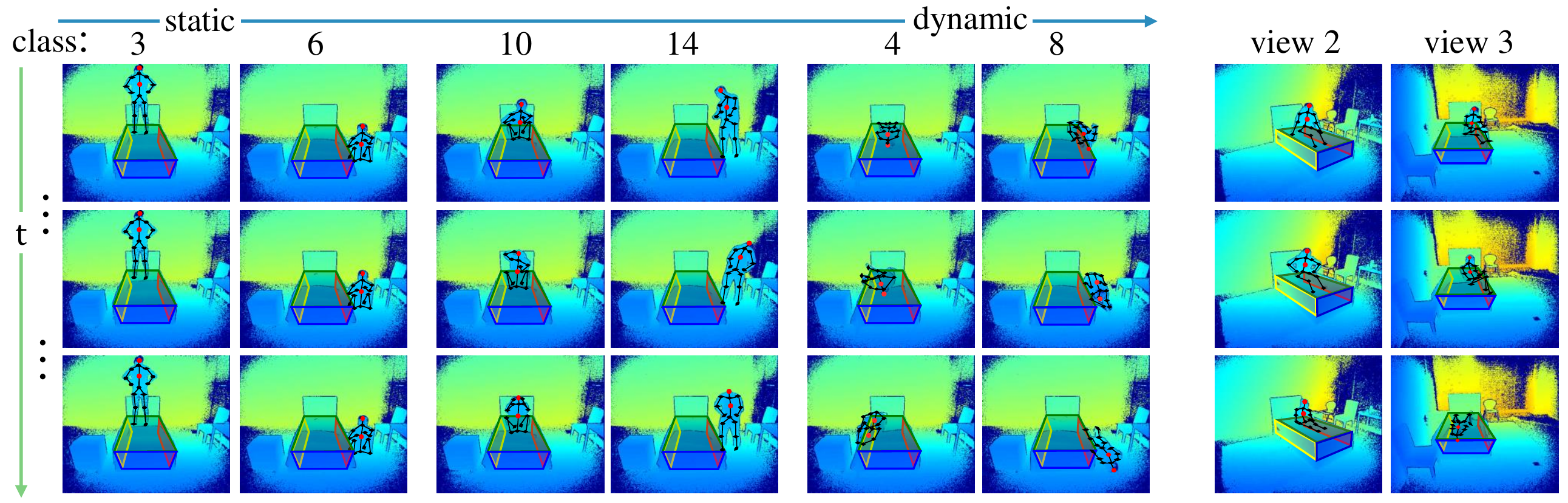}
 
\caption{Examples of our PATIENT dataset. Samples from static actions (left side) to dynamic actions (right side) are listed. Action labels are given in dataset paragraph of Sec.~\ref{sec:exper}. Examples for different views are also shown in last two columns.}
\label{fig:our_dataset}
\end{figure*}

As in Fig. \ref{fig:our_model}, variables $\alpha$ and $\beta$ first select $Q_s$ to cluster data samples according to the usefulness of the kinematic-layout $\mathcal{K}$. Then, either $Q_c$ or $Q_k$ is statistically selected to perform classification. $\alpha$, $\beta$ are set for each node as:
\begin{displaymath}
\alpha = \left\{ \begin{array}{ll}
%1 & \textrm{, if $1-\eta < \Delta < \eta$}\\
1 & \textrm{, if $|\mathcal{D}| > \eta$}\\
0 & \textrm{, otherwise}
\end{array} \right.,
\beta = \left\{ \begin{array}{ll}
1 & \textrm{, if $\zeta>\Delta$}\\
0 & \textrm{, otherwise}
\end{array} \right.
\end{displaymath}
where $|\mathcal{D}|$ is the number of samples in a current node, $\eta$ is set to $0.1$ times total number of training samples, $\Delta\in[0,1]$ is the ratio of samples having positive usefulness score $U(V)$ (Eq. \ref{useful}) in a current node and $\zeta\in[0,1]$ is a random value. %The parameter $\alpha$ has either $0$ or $1$ according to the $\Delta$ values:	
%The parameter $\eta$ is set to $0.9$ in our experiments. The variable $\beta$ is determined as follows:
%where $\zeta$ is a scalar value sampled from the uniform distribution $[0,1]$. 
As a result, $Q_s$ is first performed by $\alpha$, until a certain number of samples remain. Then, either $Q_c$ or $Q_k$ is performed by $\beta$. $1-\Delta$ is the probability for selecting $Q_c$ while $\Delta$ is the probability for selecting $Q_k$. Thus, if $\Delta$ is high, $Q_k$ is preferred to $Q_c$ while $Q_c$ is preferred to $Q_k$, otherwise. Each quality function is detailed as follows:\\
%\begin{align}
% Q_p(\Psi;\mathcal{Y}) &= \sum_{v\in\{L,R\}} \sum_{y\in\mathcal{Y}} n(y,\mathcal{D}_v) \mbox{log}\frac{n(y,\mathcal{D}_v)}{|\mathcal{D}_v|},\\
% Q_c(\Psi;\mathcal{Y}) &= \sum_{v\in\{L,R\}} \sum_{y\in\mathcal{Y}} n(y,\mathcal{D}_v) \mbox{log}\frac{n(y,\mathcal{D}_v)}{|\mathcal{D}_v|},\\
% Q_v(\Psi;\mathcal{K}) &= -\sum_{v\in\{L,R\}}\sum_{V\in \mathcal{D}_v} ||\mathcal{K}(V)-\frac{1}{|\mathcal{D}_v|}\sum_{I'\in\mathcal{D}_v} \mathcal{K}(I')||_2,\\
% Q_k(\Psi;\mathcal{Y},\mathcal{K}) &= -\sum_{v\in\{L,R\}}\sum_{V\in \mathcal{D}_v} \alpha(I)\cdot||\mathcal{F}_{\mathcal{K}}(I)-\mathcal{F}_{\mathcal{A}}(I)||_2 
%\end{align}
%where
%. We define three quality functions $\mathcal{Q}=\{Q_c, Q_v, Q_k\}$: $Q_c$ is a standard information gain for classification, $Q_v$ is a regression term on the kinematic-layout information and $Q_k$ is a classification term on the posterior of kinematic-layout information. $Q_v$ and $Q_k$ are designed to utilize kinematic-layout information prior $\mathcal{K}(V)$ when splitting the node. One of the quality functions is randomly selected at each node split as detailed below:
%without explicit extraction of it at the testing stage. 
%, as in Fig.~\ref{fig:subfig1split}:
%\begin{itemize}
\noindent \textbf{Switching term $Q_s$:} This term measures the usefulness of kinematic-layout $\mathcal{K}$ for classifying each data sample and categorizing them into two groups, where a group whose kinematic-layout is useful and another group whose kinematic-layout is less useful. The kinematic-layout $\mathcal{K}$ does not always help improve the classification task%. For example, classifying some data samples becomes much easier than before when kinematic-layout information is available, while classifying others may not
. For some samples, the appearance $\mathcal{A}$ is better (see Fig \ref{fig:uscore}). The $Q_s$ is defined to make left and right child nodes have compact usefulness scores $U(V)$ as follows:
\begin{eqnarray}
Q_s =&\bigg[ 1+
\sum_{m\in\{l,r\}}\frac{|\mathcal{D}_m|}{|\mathcal{D}|} \mbox{var}\bigg(\big\{U(V)|V\in\mathcal{D}_m\big\}\bigg)\bigg]^{-1}
\end{eqnarray}
where var($\cdot$) is the variance operator and the usefulness score $U(V)\in[-1,1]$ for sample $V$ is defined as:
\begin{eqnarray}
\label{useful}
U(V)=\mathcal{F}_{\mathcal{K},y^*}(V)-\mathcal{F}_{\mathcal{A},y^*}(V)
\end{eqnarray}
where 
$\mathcal{F}_{\mathcal{K},y^*}(V)$ and $\mathcal{F}_{\mathcal{A},y^*}(V)$ are the posterior probability for the ground-truth class label $y^*$, obtained by two pre-trained forests $\mathcal{F}_{\mathcal{K}}$ and $\mathcal{F}_{\mathcal{A}}$ by out-of-bag (OOB) samples and their kinematic-layout and appearance features respectively. If $U(V)>0$, we trust and exploit the kinematic-layout $\mathcal{K}$ \ie. ${Q_k}$ than the appearance. Otherwise, we consider that the appearance $\mathcal{A}$ is sufficient, \ie. using ${Q_c}$.\\
\noindent \textbf{Appearance term ${Q_c}$:} This term measures the {\it uncertainty} of class distributions in $\mathcal{D}_l$ and $\mathcal{D}_r$ based on the appearance $\mathcal{A}$. Standard Shannon entropy is used to evaluate the split quality as:
\begin{eqnarray}
Q_c &= \sum_{m\in\{l,r\}}|\mathcal{D}_m| \sum_{y\in\mathcal{Y}} \bigg[P(y|\{\mathcal{A}(V)|V\in\mathcal{D}_m\})\nonumber \\
&\cdot\log P(y|\{\mathcal{A}(V)|V\in\mathcal{D}_m\})\bigg],
%Q_c = \sum_{m\in\{l,r\}}|\mathcal{D}_m| \sum_{y\in\mathcal{Y}} \bigg[\frac{n(y,\mathcal{D}_m)}{|\mathcal{D}_m|}\log \frac{n(y,\mathcal{D}_m)}{|\mathcal{D}_m|}\bigg]
\end{eqnarray}
%where $n(y,\mathcal{D}_m)$ is the number of samples in $\mathcal{D}_m$ having label $y$.
This term tends to select $\Psi$ such that the class posterior distributions, empirically the class histograms, in $\mathcal{D}_l$ and $\mathcal{D}_r$ are dominated by a certain class.\\
%\noindent \textbf{Appearance term ${Q_c}$:} This term measures the {\it certainty} of class %distribution in $\mathcal{D}_l$ and $\mathcal{D}_r$ based on the appearance space %$\mathcal{A}$. The {\it entropy measure} is employed to evaluate the split quality as in %standard RFs:
%\begin{eqnarray}
%Q_c &= \sum_{m\in\{l,r\}} \sum_{y\in\mathcal{Y}} %n(y,\mathcal{D}_m)\log\frac{n(y,\mathcal{D}_m)}{|\mathcal{D}_m|}
%\end{eqnarray}
%where $n(y,\mathcal{D}_m)$ is the number of samples having class label $y$ in $\mathcal{D}_m$.
%This term selects $\Psi$ whose class histograms in two child nodes: $n(y,\mathcal{D}_l)$ and %$n(y,\mathcal{D}_r)$ have their peak.\\
\noindent \textbf{Kinematic-layout term ${Q_k}$:} This term is designed so that the kinematic-layout information $\mathcal{K}$ is expoited for training, but not needed at testing. Since $\mathcal{A}(V)$ and $\mathcal{K}(V)$ are in different feature spaces, there is a discrepancy between the two class distributions at a node: the kinematic-layout-based $P_{\mathcal{F}}(y|\{\mathcal{K}(V)|V\in\mathcal{D}\})$ and the appearance-based  $P_{\mathcal{F}}(y|\{\mathcal{A}(V)|V\in\mathcal{D}\})$, as follows:
\begin{eqnarray}
P_{\mathcal{F}}(y|\{\mathcal{A}(V)|V\in\mathcal{D}\})=\frac{1}{|\mathcal{D}|}\sum_{V\in\mathcal{D}} \mathcal{F}_{\mathcal{A}}(V), %\mathbb{I}(y|\mathcal{A}(V)) 
\label{postA}\\
P_{\mathcal{F}}(y|\{\mathcal{K}(V)|V\in\mathcal{D}\})=\frac{1}{|\mathcal{D}|}\sum_{V\in\mathcal{D}} \mathcal{F}_{\mathcal{K}}(V).
\label{postK}
\end{eqnarray}
where $\mathcal{F}_{\mathcal{A}}(V)$ and $\mathcal{F}_{\mathcal{K}}(V)$ are the posterior class distributions obtained by applying the sample $V$ to the pre-trained forests $\mathcal{F}_{\mathcal{A}}$ and $\mathcal{F}_{\mathcal{K}}$, respectively. We first reduce the gap between the distributions in order to implicitly use the kinematic-layout $\mathcal{K}$. The gap is minimized by closing the least square distance, using a weight vector $\mathbf{w^*}=[w_1,...,w_{|\mathcal{D}|}]^{\top}\in \mathbb{R}^{|\mathcal{D}| \times 1}$ as follows:
%It is multiplied to the matrix $\mathbf{A}\in \mathbb{R}^{C\times |\mathcal{D}|}$ to make their multiplication similar to the vector $\maƒhbf{b}\in\mathbb{R}^{C\times 1}$ as follows:
\begin{eqnarray}
\label{obj_w}
\mathbf{w}^* = \min_{\mathbf{w}} ||\mathbf{A}\cdot \mathbf{w} - \mathbf{b}||^2
\label{optimizeW}
 \end{eqnarray} 
\begin{algorithm}[b]
\SetAlgoLined
$\mathbf{Input}$: $\mathcal{A}(V) (\forall V\in\mathcal{D}$), $\mathcal{F}_\mathcal{A}$, $\mathcal{F}_\mathcal{K}$.\\
$\mathbf{Output}$: $\Psi^{*}$, $\mathcal{D}_l$, $\mathcal{D}_r$.\\

				\If {$\mathcal{Q}=Q_k$} 
				{
					Find $\mathbf{A}$ using
					$\mathcal{F}_{\mathcal{A}}(V)$.\\ %$\mathbb{I}(y|\mathcal{A}(V))$.\\
					Find $\mathbf{b}$ using $\mathcal{F}_{\mathcal{K}}(V)$.\\
					Find $\mathbf{w}^*$ by solving Eq. \ref{optimizeW}.\\
					Find $\Psi^*$ by minimizing Eq. \ref{Qkk}.\\
				}
				$\mathbf{if}$ {(no Information Gain or $|\mathcal{D}|\le1$)  $\mathbf{then}$ {\it Make Leaf}.} \\
				 $\mathbf{else}$ Split $\mathcal{D}$ into $\mathcal{D}_l$ and $\mathcal{D}_r$ using $\Psi^{*}$.
\caption{\label{alg:weight}Splitting a node with $Q_k$}
\end{algorithm}
%if $P(c=C(I)|\{\mathcal{A}(V)|V\in\mathcal{D}\}) < P(c=C(I)|\{\mathcal{K}(V)|V\in\mathcal{D}\})$ while becomes lower otherwise, where $C(I)$ denotes $V$'s ground-truth label.  
where $w_i$ denotes each sample's weight, the $i$-th row of $\mathbf{A}\in \mathbb{R}^{|\mathcal{Y}| \times |\mathcal{D}|}$, $A_i\in\mathbb{R}^{|\mathcal{Y}|\times 1}$ corresponds to each sample's appearance-based class distribution and each entry of $\mathbf{b}\in\mathbb{R}^{|\mathcal{Y}|\times 1}$ corresponds to the kinematic-layout-based class distribution. The closed form solution for $\mathbf{w}^*$ is obtained by multiplying the pseudoinverse of $\mathbf{A}$ to $\mathbf{b}$. Meanwhile, the weight for each sample becomes larger if its class is more important in $\mathcal{K}$ space than in $\mathcal{A}$ space, and smaller otherwise, as in Fig. \ref{fig:treegrow}. Samples with high discrepancy are highlighted and more carefully classified. The term $Q_k$ is defined on the weighted class histograms $n_{\mathbf{w}}(y,\mathcal{D}_m)$ as:
\begin{eqnarray}
Q_k &= \sum_{m\in\{l,r\}} \sum_{y\in\mathcal{Y}} n_{\mathbf{w}}(y,\mathcal{D}_m)\log\frac{n_{\mathbf{w}}(y,\mathcal{D}_m)}{\sum_{i=1}^{|\mathcal{D}|}w_i}
\label{Qkk}
\end{eqnarray}
where  $n_{\mathbf{w}}(y,\mathcal{D})=\sum_{V\in\mathcal{D}} w_i\cdot \mathbb{I}(y^*=y)$ and $\mathbb{I}(\cdot)$ is an impusle function and $y^*$ is the ground-truth class label. The overall process is summarized in Algorithm \ref{alg:weight} and $\mathbf{w}$ is used when $Q_k$ is chosen, or uniform weights are used for others. 
%\end{itemize}

\subsection{Inference by kinematic-layout-aware forests}
\label{sec:test}
At a testing stage, $\mathcal{A}(V)$ is passed down the KLRFs $\mathcal{F}^+$ by learned split functions $\{\Psi(\mathcal{A}^{\gamma}(\cdot),\tau)\}$ until it reaches the leaf nodes, which store both the class distribution $P(y|V)$ and the kinematic vectors $\mathcal{K}(V)$. The trees' responses are averaged to output the final $P(y|V)$ and $\hat{\mathcal{K}}(V)$ for each $V$.
% Both information can be further used to facilitate consistency on the information among frames for better estimating the final class label c.

%To effectively refine a forest $\mathcal{F}$ with two heterogeneous information ({\it \ie skeleton and appearance}), we utilized the global RF refinement framework \cite{GRF_CVPR2015} at the last stage. 

%In this framework, the ideal global loss is employed to refine leaf node's probabilities of trees $\mathcal{T}_t \in \mathcal{F}$:
%\begin{eqnarray}
%l(y_i,\hat{y_i}) = l\bigg(y_i,\frac{1}{T}\sum_{t=1}^T \mathcal{T}_t(\mathbf{x}_i)\bigg)
%\end{eqnarray}
%where $y_i$, $\hat{y_i}$ $\mathbf{x}_i$, $\mathcal{T}_t(\cdot)$, $T$ denotes the ground-truth label, predicted label, appearance of a sample, a response from the $t$-th tree and the number of trees respectively.

\subsection{Cross-view setting}
\label{sec:cross}
Cross-view setting is challenging: the model is testified for unseen camera views, which have much impact on the depth appearance \cite{novelview,HOPC2}. We used the CNN architecture  \cite{novelview}, which was pre-trained on multiple views. The depth-appearance $\mathcal{A}(V)$ obtained using the CNN is view-invariant to a certain degree. To further help, we augment the depth maps by synthetic rotations and translations as in \cite{wang2015cnn}, and propose one more quality function $Q_v$ in our KLRFs. We randomly switch between $Q_v$ and the combined quality function in Eq. \ref{qs}, while growing the trees.\\ 
\noindent \textbf{View clustering term ${Q_v}$:} This term measures the {\it compactness} of the data clusters in $\mathcal{D}_l$ and $\mathcal{D}_r$ using $\mathcal{K}(V)$:
\begin{eqnarray}
Q_v=&\bigg[ 1+ \sum_{m\in\{l,r\}}\frac{|\mathcal{D}_m|}{|\mathcal{D}|}\Lambda\bigg(\big\{\mathcal{K}(V)|V\in\mathcal{D}_m\big\}\bigg) \bigg]^{-1}
\end{eqnarray}
where $\Lambda=\mbox{trace(var(}\cdot\mbox{))}$ is defined as trace of a variance operator. Since the augmented data shares the same kinematic-layout information, they are clustered together and this further helps deal with the view-invariance. Since the kinematic-layout information and action classes are correlated \cite{Fouhey2014,Fouhey_2012,CVPR2013_daction2}, $Q_v$ further facilitates data separation by their action labels.\\
\noindent \textbf{Exploiting kinematic consistency.}
After obtaining both $P(y|V)$ and  $\mathcal{\hat{K}}(V)$ from the leaf nodes, we reduce noises in their responses by applying the kinematic consistency filter (KCF) to $P(y|V)$, exploiting pairwise similarities of inferred $\mathcal{\hat{K}}(V)$ as follows:
\begin{eqnarray}
P^{*}(y|V)=\frac{1}{W_p}\sum_{J'\in \mathcal{S}(V)} P(y|J) g(||\hat{\mathcal{K}}(J)-\hat{\mathcal{K}}(J')||) 
\end{eqnarray}
where $W_p=\sum_{ J'\in\mathcal{S}(V)}g(||\hat{\mathcal{K}}(J)-\hat{\mathcal{K}}(J')||)$ is a normalizing factor, $g(\cdot)$ is a Gaussian kernel and $S(V)$ is the augmented data set of $V$.
%The posterior of action class probability given a depth sequence $V$ is defined as: $P(-E(y|V)) = \exp{\{-\sum_{t=1}^T \phi_{u}(I_t)  + \sum_{t < t'} \phi_{p} (I_t, I_{t'} )\}}$, where the unary potential $\phi_u(I_t) =\log \\ \big(P(y|\mathcal{A}(I_t))\big)$ is defined as each frame's probability that is obtained from the forest $\mathcal{F}$ in Sec \ref{sec:forest} and the pairwise potentials in our model have the form:
%\begin{eqnarray}
%\phi_{p} (I_t, I_{t'} ) &=& \sigma \cdot \mathbf{1}(c_t\ne c_{t'}) \cdot\exp\bigg(-\frac{|t-{t'}|^2}{2\theta_{\alpha}^2} \nonumber \\ 
%&&-\frac{|\mathcal{\hat{K}}(I_t)-\mathcal{\hat{K}}(I_{t'})|^2}{2\theta_{\beta}^2}\bigg)
%\end{eqnarray}
%, where $\mathbf{1}(\cdot)$ denotes the indicator function. The label compatibility term $\mathbf{1}(c_t\ne c_{t'})$, given by the Potts model, introduces a penalty to temporally near frames that are assigned with different labels. The overall pair-wise potential is designed to make temporally near frames with similarly inferred skeletal information share their class information. The degrees of similarity is controlled by parameters $\theta_{\alpha}$, $\theta_{\beta}$ and $\sigma$ are linear combination weights and obtained empirically. We find the refined class configurations for each frame by minimizing the energy function $E(y|V)$ with mean-field approximation \cite{pk_nips_2011}.

\section{Experiments}
\label{sec:exper}

\begin{table}[t]
\footnotesize
\setlength{\tabcolsep}{1.2em}
\centering
\scalebox{1.4}{\begin{tabular}{ l | c }
\Xhline{1pt}
 \multirow{1}{*}{Method} & \multirow{1}{*}{View 1 (\%)} \\ 
 \hline
\hline
DCSF \cite{Lu_cvpr_2013}  &  $18.7$  \\ 
\hline
HON4D \cite{Omar_cvpr_2013}  & $21.1$ \\ 
\hline
HOPC \cite{HOPC2}  & $28.2$  \\ 
\hline
DMM \cite{wang2015cnn}  & $29.3$  \\ 
\hline
Novel View \cite{novelview} & $43.8$ \\
%\hline
%\multicolumn{2}{c}{$\mathcal{A}(V)$: Depth}\\
\hline
\hline
Baseline (RFs) &   $47.8$    \\ 
\hline
Ours (KLRFs) & $\mathbf{53.2}$   \\ 
\Xhline{1pt}
\end{tabular}}
\caption{Performance on PATIENT (same-view).}
\label{tab:ourPATIENTsame}
\end{table}

We perform both single-view (on PATIENT, CAD-60 \cite{sung_rgbdactivity_2012} datasets) and cross-view (on PATIENT, UWA3D Multiview Activity II \cite{HOPC2} datasets) experiments to validate our methods. Accuracy averaged for $10$ trials are reported in each table. Experimental settings and results are as follows:\\
\noindent \textbf{Datasets.} In \textbf{PATIENT} dataset, $10$ different subjects perform $15$ different actions with $3$ different views, whose actions are mainly defined by their interactions to the {\it bed} and {\it floor} planes. As in Fig. \ref{fig:our_dataset}, our dataset contains both static and dynamic actions and all $15$ actions are: (1) lying, (2) sitting and (3) standing on the bed; (4-5) stretching body parts out of the bed when the patient is lying and sitting; (6-7) sitting and standing on the floor; (8) falling out of the bed; (9-15) suffering status of actions(1-8) except (3). 
%Overall, there are $15$ distinct classes in PATIENT action dataset.
In \textbf{CAD60} dataset \cite{sung_rgbdactivity_2012}, there are $68$ video clips including RGB, depth and skeleton joints. There are $4$ different subjects performing $14$ different actions: (1) still, (2) talking on the phone, (3) writing on white board, (4) drinking water, (5) rinsing mouth with water, (6) brushing teeth, (7) wearing contact lenses, (8) talking on couch, (9) relaxing on couch, (10) chopping, (11) stirring, (12) opening pill container, (13) working on computer, (14) random. These actions are completed in 5 different indoor environments: ({\it office, kitchen, bedroom, bathroom, and living room}). In \textbf{UWA3D Multiview} dataset \cite{HOPC2}, there are $1075$ video clips including RGB, depth and skeleton joints. There are $10$ different subjects, performing $30$ different actions with $4$ different camera views: (1) one hand waving, (2) one hand punching, (3) two hand waving, (4) two hand punching, (5) sitting down, (6) standing up, (7) vibrating, (8) falling down, (9) holding chest, (10) holding head, (11) holding back, (12) walking, (13) irregular walking, (14) lying down, (15) turning around, (16) drinking, (17) phone answering, (18) bending, (19) jumping jack, (20) running, (21) picking up, (22) putting down, (23) kicking, (24) jumping, (25) dancing, (26) moping ﬂoor, (27) sneezing, (28) sitting down (chair), (29) squatting, and (30) coughing.\\
\noindent \textbf{Depth appearance $\mathcal{A}$.} We resize each depth map into $256 \times 256$ first, then apply $10$ times translation augmentation to generate $227 \times 227$ maps, as in \cite{Alex_nips_2012} and additionally we applied $5$ times random rotation (\ie each frame is 3D rotated with random angles between $0$ to $60$ degrees) and $10$ times temporal augmentation (\ie $10$ offsets are used when doing the fourier transform), similar to \cite{wang2015cnn}.\\
\noindent \textbf{Kinematic-layout $\mathcal{K}$.} Kinematic-layout information $\mathcal{K}$ is extracted based on both layout planes $\mathbb{L}$ and human skeleton joints $\mathbb{P}(t)$, as defined in Sec. \ref{sec:priors}. Since the information is different across datasets, we defined different configurations for $\mathbb{L}$ and $\mathbb{P}(t)$, depending on the usefulness of the information. For PATIENT dataset, human skeleton joints cannot be easily obtained due to challenging viewpoints. Thus, we manually labeled {\it head} and {\it body} positions to construct $\mathbb{P}(t)$. For the $\mathbb{L}$, we generated 5 planes: {\it bed top}, {\it bed left}, {\it bed right}, {\it bed front} and {\it floor} for each sequence. For CAD60 and UWA3D Multiview Activity II datasets, we use the available $15$ human skeleton joints for $\mathbb{P}(t)$ and manually labeled $5$ layout planes $\mathbb{L}$: {\it ceiling}, {\it floor}, {\it left wall}, {\it middle wall}, {\it right wall} for each sequence.

\begin{table}[t]

\footnotesize
\setlength{\tabcolsep}{1.1em}
\centering

%\begin{tabular}{l || l | l | l ||  l  || l  }
%\resizebox{0.8\columnwidth}{!}{
\scalebox{1}{\begin{tabular}{l |  c c c   }

\Xhline{1pt}
%\tabucline[2pt]{-}
%\tabucline[2pt]{-}
%\multicolumn{5}{c}{$\mathcal{A}(V)$: Depth}\\
%\Xhline{1.3pt}
 %& \multicolumn{3}{c|}{aPascal \cite{Farhadi09CVPR}} & \multicolumn{3}{c}{ImageNet \cite{RussakovskyECCV10}}  \\

%\cline{2-7}
%&  AP@2  &  AP@5  &  AP@8   &  AP@2  &  AP@5  &  AP@8  \\
Method & Accuracy  & Precision & Recall \\ 
\Xhline{1pt}
\multicolumn{4}{c}{Testing Input: Depth}\\
\hline
HON4D \cite{Omar_cvpr_2013}  &    72.7 & $-$ & $-$ \\ 
\hline
Zhu \etal  \cite{CVPR2013_daction2} & 75.0  & $-$ & $-$ \\ 
\hline
\hline
Baseline (RFs)  & 81.6 & 93.2 & 78.6\\ 
\hline
Ours (KLRFs)  & 87.1 & 92.3 & 85.7 \\ 
%Ours (KLRF+CRF) &    87.3 & &\\ 
\Xhline{1pt}
%\tabucline[2pt]{-}
%\tabucline[2pt]{-}
%\hline
%\hline
\multicolumn{4}{c}{Testing Input: RGB+Skeleton}\\
\hline
STIP \cite{Zhu_2014_IVC} &  62.5  & $-$&$-$\\ 
\hline
order sparse coding \cite{Bingbing_eccv_2012} &  65.3  & $-$ & $-$ \\ 
\hline
\multicolumn{4}{c}{Testing Input: RGB+Depth+Skeleton}\\
\hline
object affordance \cite{Koppula_2013}  &  71.4 &$-$ &$-$\\ 
\hline
JOULE \cite{CVPR15_heterogeneous}  &  84.1 & $-$ & $-$\\ 
\hline
\multicolumn{4}{c}{Testing Input: Skeleton}\\
\hline
GI \etal \cite{CADstate2} &   $-$ & $91.9$ & $90.2$\\ 
\hline
Shan \etal \cite{CADstate3} &   $91.9$ & $93.8$ & $\mathbf{94.5}$\\ 
\hline
Cippitelli \etal \cite{CADstate1} &  $-$ & $93.9$ & $93.5$\\
\hline
\multicolumn{4}{c}{Testing Input: Depth+Skeleton}\\
\hline
Actionlet Ensemble \cite{Jiang_tpami_2014} &   74.7 & $-$ & $-$\\ 
\hline
Zhu \etal \cite{CVPR2013_daction2} &   $87.5$ & $93.2$ & $84.6$\\ 
%Novel View \cite{novelview} &   \checkmark &     &  & \\
%\hline
%\hline
%\seungryul{Depth/RGB/Skeleton-based RF} & \checkmark  & \checkmark & \checkmark & \checkmark & \textbf{} & & \\ 
%\hline
%\seungryul{Depth/Skeleton-based RF} &  & \checkmark & \checkmark & \checkmark &  87.5 & & \\ 
%\hline
\hline\hline
Baseline (RFs+Skeleton) &  $89.7$ & $92.9$ & $89.3$\\
\hline
Ours (KLRFs+Skeleton) & $\mathbf{94.1}$ & $\mathbf{97.5}$ & $92.7$\\ 

%Ours (KLRF+CRF) &  &  & \\
%Ours+RGB/Skeleton-based RF & \checkmark  & \checkmark & \checkmark & \checkmark & \textbf{} & & \\ 
\Xhline{1pt}
\end{tabular}}
\caption{Performance on CAD-60 (same-view)}
\label{tab:ourCAD}
\end{table}

\subsection{Same-view action recognition}
We first evaluate our method for single-view action recognition using PATIENT, CAD60 datasets.\\
\noindent \textbf{Results on PATIENT dataset.} The results of the PATIENT dataset is presented in the Table \ref{tab:ourPATIENTsame}. The classification accuracy is averaged over all classes, which corresponds to the mean of the confusion matrix diagonal. We use the first $5$ subjects as training and others as testing samples. We evaluate the recent state-of-the-art depth-based methods \cite{Lu_cvpr_2013,Omar_cvpr_2013,HOPC2,wang2015cnn,novelview} on our PATIENT dataset using their publicly available codes. Our method produces a significant performance ($6-10\%$) gain over these methods. 
%\vspace{-0.1cm}
\begin{table}[t]

%\small
%\begin{tabular*}{5cm}{@{\extracolsep{\fill}}lllr}

%\footnotesize
%\begin{tabular}{l || l | l | l ||  l  || l  }
%\resizebox{0.8\columnwidth}{!}{
%\begin{minipage}{\textwidth}
\footnotesize
\setlength{\tabcolsep}{0.7em}
\centering
\scalebox{1.3}{\begin{tabular}{ l || c | c || c}
\Xhline{1pt}

 %& \multicolumn{3}{c|}{aPascal \cite{Farhadi09CVPR}} & \multicolumn{3}{c}{ImageNet \cite{RussakovskyECCV10}}  \\

%\cline{2-7}
%&  AP@2  &  AP@5  &  AP@8   &  AP@2  &  AP@5  &  AP@8  \\

 \multirow{2}{*}{Method} & \multicolumn{2}{c||}{PATIENT} &  UWA3D\\
\cline{2-3}
%\hhline{~--~}
& View 2   & View 3 &  Multiview\\
 \hline
\hline
DCSF \cite{Lu_cvpr_2013}  &  $6.7$ & $16.0$  &  $-$ \\ 
\hline
HON4D \cite{Omar_cvpr_2013}  & $6.3$ & $13.8$  & $28.9$\\ 
\hline
HOPC \cite{HOPC2}  & $15.4$   & $23.1$ & $52.2$\\ 
\hline
DMM \cite{wang2015cnn}  & $19.3$   & $24.0$& $-$ \\ 
\hline
Novel View \cite{novelview} & $23.8$ & $32.5$ & $76.9$\\
\hline
%\hline
%\multicolumn{3}{c}{$\mathcal{A}(V)$: Depth}\\
\hline
Baseline (RFs) &   $21.5$   & $27.2$ & $77.1$\\ 
\hline
Ours (KLRFs) & $\mathbf{27.5}$   & $\mathbf{36.2}$ & $\mathbf{80.4}$\\ 
\Xhline{1pt}
%Ours (KLRF+CRF) & & & \\
%\seungryul{Ours w/ HOPC} \cite{HOPC2} &   $\textbf{}$ & \\

\end{tabular}}

\caption{Performance on both PATIENT and UWA3D Multiview Activity II (cross-view).}
\label{tab:crossview}
\end{table}

%\vspace{-0.1cm}

\noindent \textbf{Results on CAD60 dataset.} For the CAD60, we follow the cross-person experimental setting of \cite{sung_rgbdactivity_2012} and use $3$ measures (\ie accuracy \cite{CVPR15_heterogeneous,Jiang_tpami_2014}, precision/recall \cite{sung_rgbdactivity_2012}) to compare with many state-of-the-arts. The result is shown in Table \ref{tab:ourCAD}. We obtained good performance using only depths as inputs (denoted as KLRFs in  the top of Table \ref{tab:ourCAD}), which is also comparable to other methods using different input cues (\eg additional RGB and skeletons). Since this dataset contains mostly frontal humans with frontal camera views where skeleton estimation performs well, most state-of-the-arts use skeleton joints to obtain their results. We also report the best results by combining skeletons at testing. (denoted as KLRFs+Skeleton in the bottom of Table \ref{tab:ourCAD}).

%\begin{figure}[h]
%\centering
%   \includegraphics[scale =0.4] {CADclass.eps}\hspace{10pt}
   
%\caption{Per class accuracy of the RF and KLRF method for CAD-60 dataset}
%\label{fig:cadclass}
%\end{figure}
\setlength{\belowcaptionskip}{-1pt}

\begin{figure}[b]
\centering
\subfigure[PATIENT dataset.]{
   \includegraphics[width=0.45\linewidth] {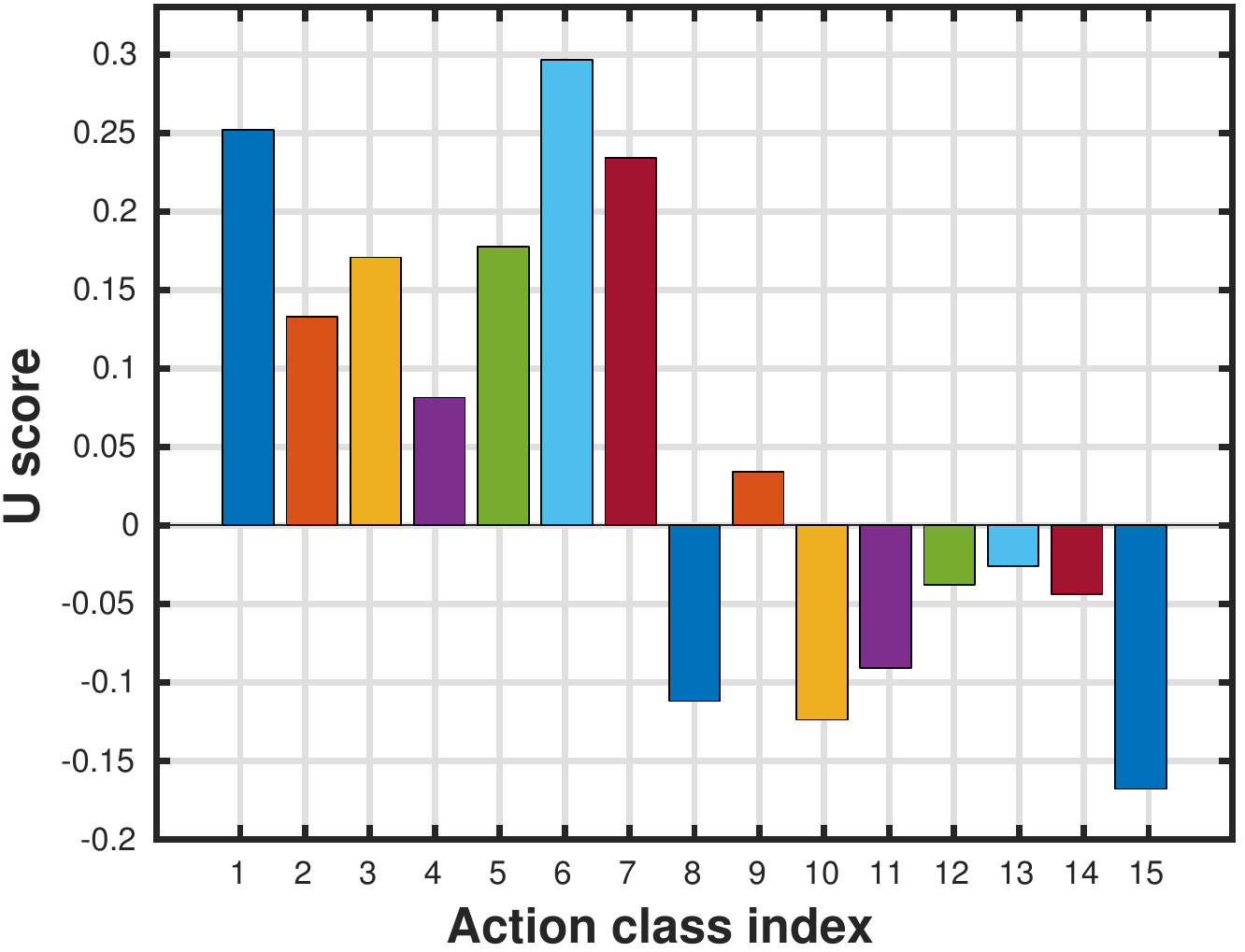}
   \label{fig:subfig1split}   
 }
 \vspace{-0.3cm}
 \subfigure[CAD-$60$ dataset.]{
   \includegraphics[width=0.45\linewidth] {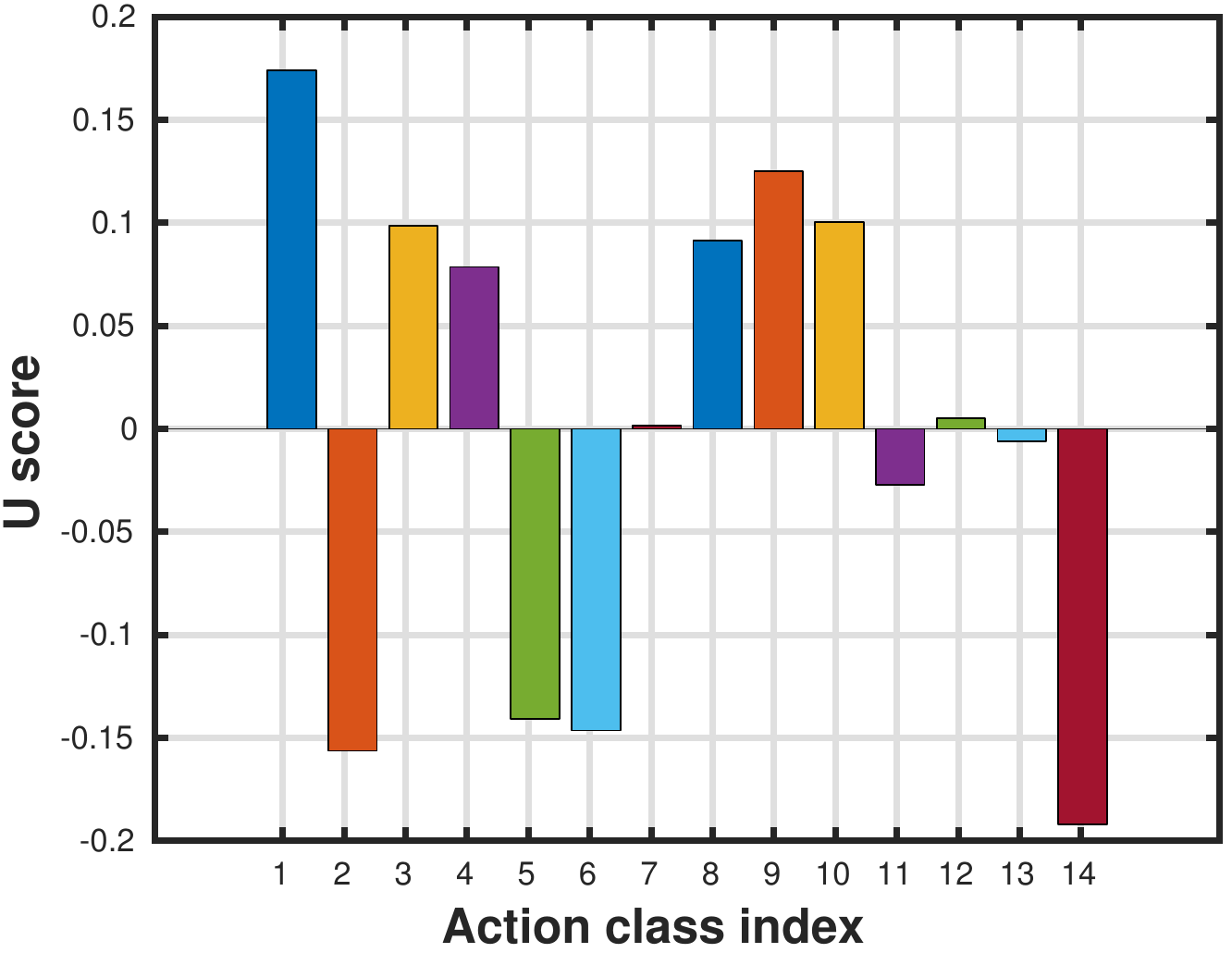}
   \label{fig:subfig1split}   
 } 
% \vspace{-0.3cm}
 \subfigure[UWA3D Multiview Activity II dataset.]{
   \includegraphics[width=0.85\linewidth] {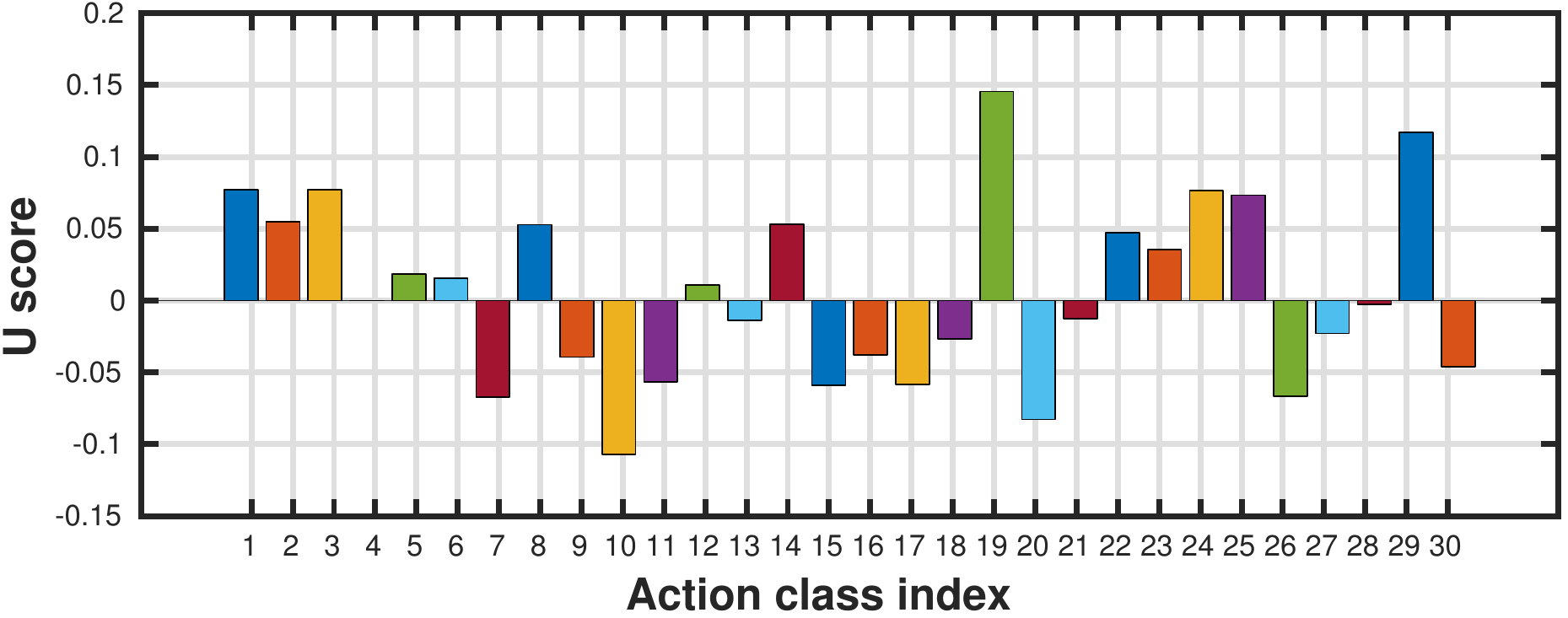}
   \label{fig:subfig1split}   
 }
\caption{Usefulness score $U(V)$ vs. class labels. Class labels are given in the dataset paragraph of Sec.~\ref{sec:exper}.}
\label{fig:uscore}
\end{figure}

\subsection{Cross-view action recognition}

We applied the same model as in the same-view experiment to test $5$ subjects in different views for the cross-view experiments. The results are summarized in the left side of the Table \ref{tab:crossview} for two different views (\ie View 2, 3) of the PATIENT dataset. We also trained models for the UWA3D Multiview dataset as in \cite{novelview} and their mean accuracies are reported in the right side of the Table \ref{tab:crossview}. Detailed results for UWA3D are given in supplementary pages.

\begin{figure}[t]
\centering
\includegraphics[scale =0.2] {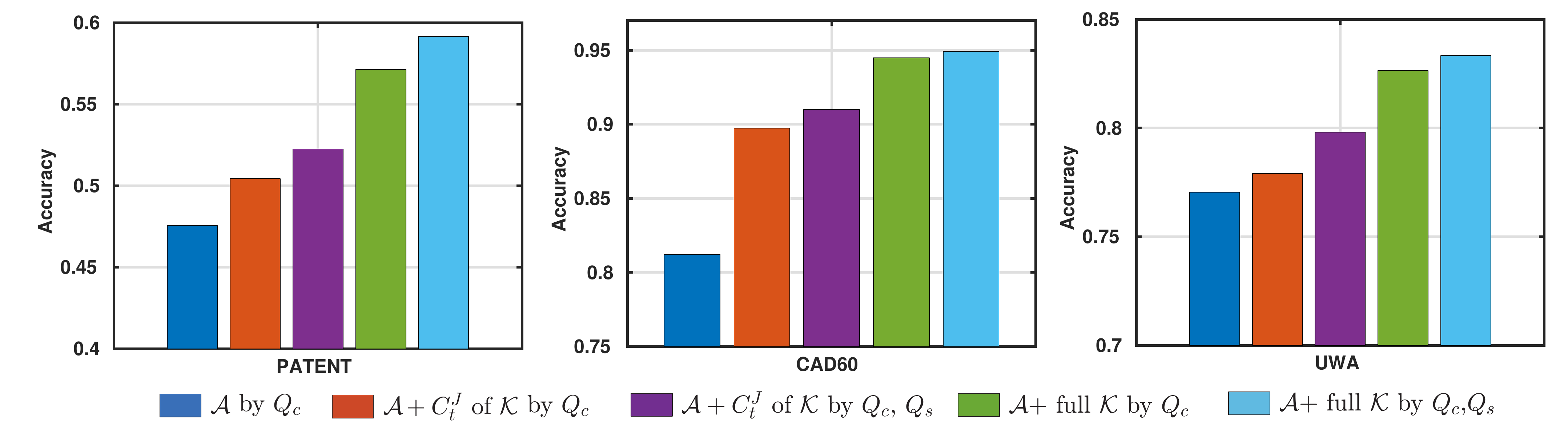}
 %  \includegraphics[scale =0.2] {bar_patent_1.pdf}
 %  \includegraphics[scale =0.2] {bar_cad_1.pdf}
 %  \includegraphics[scale =0.2] {bar_uwa_1.pdf}\\
 %  \includegraphics[scale =0.16] {bar_label1.png}
 %  \includegraphics[scale =0.16] {bar_label1.png}
 %  \includegraphics[scale =0.16] {bar_label1.png}\\
 %  \vspace{-0.1cm}
   \caption{Utilizing the $\mathcal{K}$ at testing stage.}
\label{fig:baseline}
\end{figure}
\begin{figure}[t]
\centering
\includegraphics[scale =0.2] {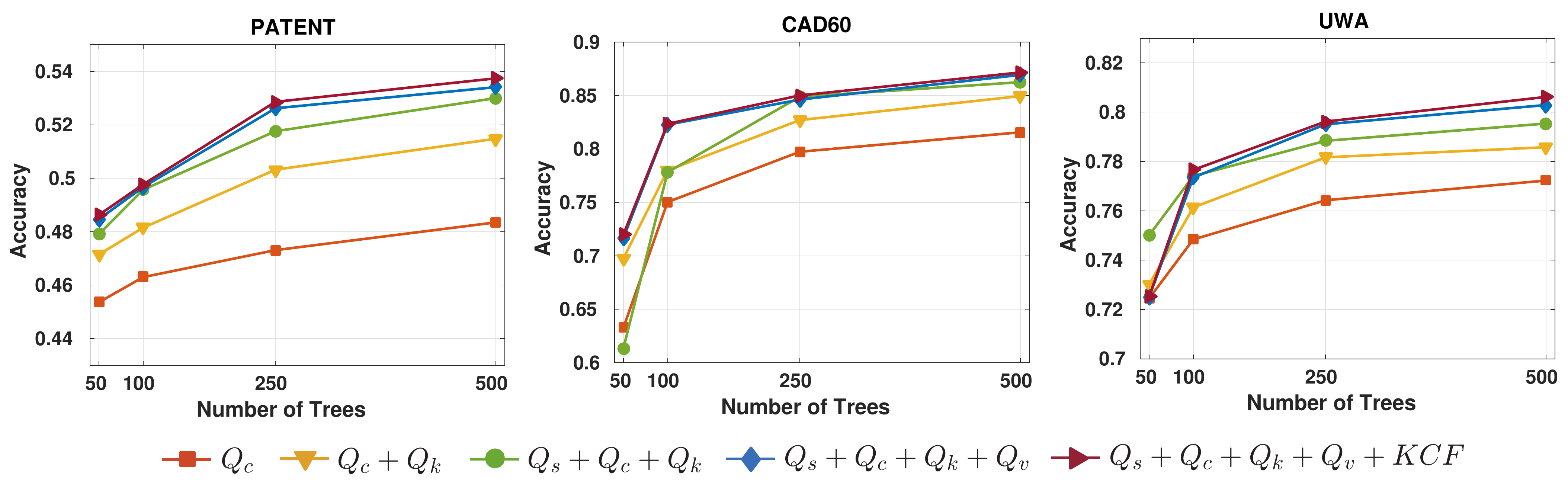}
   \caption{Sensitivity/component analysis for our KLRFs.}
   \vspace{-0.5cm}
\label{fig:component}
\end{figure}

\subsection{Further analysis}
%We analyze our method with $3$ more experiments:
%\noindent \textbf{Effect of each component.} To analyze the contribution of each component in our method, we provide experimental results on both datasets using the individual term and the combination of them. In Table \ref{tab:component}, we show the element-wise performance comparison for our algorithm. We grow forest $\mathcal{F}$ by 1) $Q_c$, 2) $Q_c$ and $Q_v$, 3) $Q_c$, $Q_v$, $Q_k$ and 4) exploiting consistency. Our method consistently improves the performance by adding each component. Exploiting the consistency with CRF provides a small improvement in PATIENT dataset. This is due to the fact that the PATIENT contains more static actions, where the temporal smoothing is not necessary in this case.
%using CRF model doesn't perform well when human actions are static. Also, we perform our experiment with static classes only (Row 1) and with both dynamic and static classes (Row 2). 

%\begin{table}[h]
%\footnotesize
%\small
%\begin{tabular*}{5cm}{@{\extracolsep{\fill}}lllr}
%\setlength{\tabcolsep}{0.5em}

%\centering
{%\footnotesize
%\begin{tabular}{l || l | l | l ||  l  || l  }
%\resizebox{0.8\columnwidth}{!}{
%\begin{tabular}{ c  | c | c | c}

 %& \multicolumn{3}{c|}{aPascal \cite{Farhadi09CVPR}} & \multicolumn{3}{c}{ImageNet \cite{RussakovskyECCV10}}  \\

%\cline{2-7}
%&  AP@2  &  AP@5  &  AP@8   &  AP@2  &  AP@5  &  AP@8  \\

%\hline
%    \multirow{1}{*}{Method} & \multirow{1}{*}{PATIENT}  & \multirow{1}{*}{CAD60} & %\multirow{1}{*}{UWA3D}\\
%\hline
%  $Q_c$ &  $50.04\pm0.75$  & $78.24\pm0.24$ & \\
%$Q_c$, $Q_v$ & $51.72\pm0.59$   & $79.12\pm0.11$ &  \\   
%$Q_c$, $Q_v$, $Q_k$, $Q_p$ & $54.86\pm0.80$ & $84.71\pm0.32$ & \\
 %$Q_c$, $Q_v$, $Q_k$ + CRF & $55.33\pm0.95$ &  $87.35\pm0.12$ &\\
%\hline

%\end{tabular}}
%\caption{Comparison of accuracy for each components on PATIENT, CAD60 and UWA3D datasets. We run the experiments $5$ times for $4$ different splits (total $20$ times) and report their average accuracy.}

%}
%\label{tab:component}
%\end{table}

\noindent \textbf{Usefulness score $U$ vs. classes.} In Fig \ref{fig:uscore}. we plot the averaged usefulness score $U(V)\in[-1,1]$ for samples in each action classes. If $U(V)>0$, $\mathcal{K}$ is regarded more useful than $\mathcal{A}$, while $\mathcal{A}$ is regarded more useful than $\mathcal{K}$, otherwise. In PATIENT dataset, the results implies that static actions as (1)-(7) are ambiguous in $\mathcal{A}$ space while $\mathcal{K}$ space explains those actions well. Dynamic actions as (8)-(15) are well classified using only $\mathcal{A}$ space. In CAD-60 and UWA3D datasets, we also report their results, showing variations.\\
\noindent \textbf{Utilizing $\mathcal{K}$ at testing stage.}  To test the strength of kinematic-layout $\mathcal{K}$, we report the performance by explicitly using ground-truths of $\mathcal{K}$ as input features in Fig. \ref{fig:baseline}. Note that utilizing $\mathcal{K}$ at testing stage is not realistic, since it is secured by ground-truths. Experiments are conducted for only evaluation purpose, configuring $3$ features $\{$ $\mathcal{A}$, $\mathcal{A}+\mathbf{C}^J_t$ of $\mathcal{K}$, $\mathcal{A}+\mathcal{K}\}$ and $2$ classifiers $\{$RFs (\ie $Q_c$), KLRFs using $Q_c$, $Q_s\}$ for each dataset. The graph shows that $\mathcal{K}$ offers $5-10\%$ accuracy gain, when combined with $\mathcal{A}$.

\noindent \textbf{Sensitivity/component analysis.} We evaluate the sensitivity of our model depending on tree numbers in Fig. \ref{fig:component}. The performance increases as tree numbers increase and saturates around $500$ trees. Thus, we set tree numbers as $500$. Component analysis is further performed by turning on/off quality functions and KCF. We report them in the same figure.

\vspace{-0.1cm}

\section{Conclusion}
\label{sec:con}

In this paper, we study the problem of action recognition in a scenario of $24$ hours-monitoring patient actions in a ward, with the goal of effectively recognizing both static actions such as ``lying on the bed'' and dynamic actions such as ``falling out of the bed''. %In order to capture subtle or significant temporal movements, we incorporate the scene layout and skeleton information in the learning process. 
We propose the kinematic-layout-aware random forest to encode the scene layout and skeleton information as privileged information, thereby capturing more geometry and kinematic-layout information   providing greater discriminative power in action classification.

{\small
\bibliographystyle{ieee}
\bibliography{egbib}

\begin{thebibliography}{10}\itemsep=-1pt

\bibitem{Cheng_eccv_2012}
Z.~Cheng, L.~Qin, Y.~Ye, Q.~Huang, and Q.~Tian.
\newblock {Human daily action analysis with multi-view and color-depth data}.
\newblock In {\em ECCV Workshop}, 2012.

\bibitem{cheronICCV15}
G.~Ch{\'e}ron, I.~Laptev, and C.~Schmid.
\newblock {P-CNN: pose-based CNN features for action recognition}.
\newblock In {\em ICCV}, 2015.

\bibitem{CADstate1}
E.~Cippitelli, S.~Gasparrini, E.~Gambi, and S.~Spinsante.
\newblock {A human activity recognition system using skeleton data from RGBD
  sensors}.
\newblock In {\em Computational Intelligence and Neuroscience}, 2016.

\bibitem{couprie_iclr_13}
C.~Couprie, C.~Farabet, L.~Najman, and Y.~LeCun.
\newblock {Indoor semantic segmentation using depth information}.
\newblock In {\em ICLR}, 2013.

\bibitem{dantone_2012_cvpr}
M.~Dantone, J.~Gall, G.~Fanelli, and L.~V. Gool.
\newblock {Real-time facial feature detection using conditional regression
  forests }.
\newblock In {\em CVPR}, 2012.

\bibitem{Dapogny_2015_ICCV}
A.~Dapogny, K.~Bailly, and S.~Dubuisson.
\newblock {Pairwise conditional random forests for facial expression
  recognition}.
\newblock In {\em ICCV}, 2015.

\bibitem{Fouhey_2012}
V.~Delaitre, D.~F. Fouhey, I.~Laptev, J.~Sivic, A.~Gupta, and A.~A. Efros.
\newblock {Scene semantics from long-term observation of people}.
\newblock In {\em ECCV}, 2012.

\bibitem{Dollar_2005}
P.~Doll\'ar, , V.~Rabaud, G.~Cottrell, and S.~Belongie.
\newblock {Behavior recognition via sparse spatio-temporal features}.
\newblock In {\em IEEE International Workshop on Visual Surveillance and
  Performance Evaluation of Tracking and Surveillance}, 2005.

\bibitem{PSSR1}
V.~Ferrari, M.~Marin-Jimenez, and A.~Zisserman.
\newblock {Progressive search space reduction for human pose estimation}.
\newblock In {\em CVPR}, 2008.

\bibitem{Fouhey2014}
D.~F. Fouhey, V.~Delaitre, A.~Gupta, A.~A. Efros, I.~Laptev, and J.~Sivic.
\newblock {People watching: human actions as a cue for single view geometry}.
\newblock {\em IJCV}, 2014.

\bibitem{Gall2010}
J.~Gall, A.~Yao, and L.~V. Gool.
\newblock {2D action recognition serves 3D human pose estimation}.
\newblock In {\em ECCV}, 2010.

\bibitem{CADstate2}
P.~GI, W.~C, and W.~S.
\newblock {Self-organizing neural integration of pose-motion features for human
  action recognition}.
\newblock In {\em Frontier in Neurobotics}, 2015.

\bibitem{Gkioxari_2015_cvpr}
G.~Gkioxari and J.~Malik.
\newblock {Finding action tubes}.
\newblock In {\em CVPR}, 2015.

\bibitem{Hedau_iccv_2009}
V.~Hedau, D.~Hoiem, and D.~Forsyth.
\newblock {Recovering the spatial layout of cluttered rooms.}
\newblock In {\em ICCV}, 2009.

\bibitem{CVPR15_heterogeneous}
J.-F. Hu, W.-S. Zheng, J.~Lai, and J.~Zhang.
\newblock {Jointly learning heterogeneous features for RGB-D activity
  recognition}.
\newblock In {\em CVPR}, 2015.

\bibitem{RFaction}
A.~Joshia, C.~Monnierb, M.~Betkea, and S.~Sclaroff.
\newblock {Comparing random forest approaches to segmenting and classifying
  gestures}.
\newblock {\em Image and Vision Computing}, 2016.

\bibitem{klaser_bmvc_2008}
A.~Klaser, M.~Marszalek, and C.~Schmid.
\newblock {A spatio-temporal descriptor based on 3D-gradients}.
\newblock In {\em BMVC}, 2008.

\bibitem{Kong_2015_cvpr}
Y.~Kong and Y.~Fu.
\newblock {Bilinear heterogeneous information machine for RGB-D action
  recognition}.
\newblock In {\em CVPR}, 2015.

\bibitem{Koppula_2013}
H.~S. Koppula, R.~Gupta, and A.~Saxena.
\newblock Learning human activities and object affordances from rgb-d videos.
\newblock {\em IJRR}, 2013.

\bibitem{Alex_nips_2012}
A.~Krizhevsky, Sutskever, Ilya, and G.~E. Hinton.
\newblock {ImageNet Classification with Deep Convolutional Neural Networks}.
\newblock In {\em NIPS}, 2012.

\bibitem{Laptev_2005_ijcv}
I.~Laptev.
\newblock {On space-time interest points}.
\newblock {\em IJCV}, 2005.

\bibitem{Laptev_2008_CVPR}
I.~Laptev, M.~Marszalek, C.~Schmid, and B.~Rozenfeld.
\newblock {Learning realistic human actions from movies}.
\newblock In {\em CVPR}, 2008.

\bibitem{Wanqing_cvprw_2010}
W.~Li, Z.~Zhang, and Z.~Liu.
\newblock {Action recognition based on a bag of 3D points}.
\newblock In {\em CVPR Workshop}, 2010.

\bibitem{Cewu_cvpr_2014}
C.~Lu, J.~Jia, and C.-K. Tang.
\newblock {Range-Sample Depth Feature for Action Recognition}.
\newblock In {\em CVPR}, 2014.

\bibitem{Jiajia_iccv_2013}
J.~Luo, W.~Wang, and H.~Qi.
\newblock {Group sparsity and geometry constrained dictionary learning for
  action recognition from depth maps}.
\newblock In {\em ICCV}, 2013.

\bibitem{Lv_cvpr_2007}
F.~Lv and R.~Nevatia.
\newblock {Single view human action recognition using key pose matching and
  viterbi path searching}.
\newblock In {\em CVPR}, 2007.

\bibitem{reglstm}
B.~Mahasseni and S.~Todorovic.
\newblock {Regularizing long short term memory with 3D human-skeleton sequences
  for action recognition}.
\newblock In {\em CVPR}, 2016.

\bibitem{Bingbing_eccv_2012}
B.~Ni, P.~Moulin, and S.~Yan.
\newblock Order-preserving sparse coding for sequence classification.
\newblock In {\em ECCV}, 2012.

\bibitem{Bingbing_2011_iccv}
B.~Ni, G.~Wang, and P.~Moulin.
\newblock {RGBD-HuDaAct: a color-depth video database for human daily activity
  recognition}.
\newblock In {\em ICCV Workshop}, 2011.

\bibitem{Nie_2015_CVPR}
B.~X. Nie, C.~Xiong, and S.-C. Zhu.
\newblock {Joint action recognition and pose estimation from video}.
\newblock In {\em CVPR}, 2015.

\bibitem{Omar_cvpr_2013}
O.~Oreifej and Z.~Liu.
\newblock {HON4D: histogram of oriented 4D normals for activity recognition
  from depth sequences}.
\newblock In {\em CVPR}, 2013.

\bibitem{Pauly_2003}
M.~Pauly, R.~Keiser, and M.~Gross.
\newblock {Multi-scale feature extraction on point-sampled surfaces}.
\newblock {\em Computer Graphics Forum}, 2003.

\bibitem{Pham_2015_ICCV}
V.-Q. Pham, T.~Kozakaya, O.~Yamaguchi, and R.~Okada.
\newblock {COUNT forest: co-voting uncertain number of targets using random
  forest for crowd density estimation}.
\newblock In {\em ICCV}, 2015.

\bibitem{hanklett}
L.~L. Presti, M.~L. Cascia, S.~Sclaroff, and O.~Camps.
\newblock {Hankelet-based dynamical systems modeling for 3D action
  recognition}.
\newblock {\em Image and Vision Computing}, 2015.

\bibitem{HOPC2}
H.~Rahmani, A.~Mahmood, D.~Q. Huynh, and A.~Mian.
\newblock {Histogram of oriented principal components for cross-view action
  recognition}.
\newblock {\em TPAMI}, 2016.

\bibitem{novelview}
H.~Rahmani and A.~Mian.
\newblock {3D action recognition from novel viewpoints}.
\newblock In {\em CVPR}, 2016.

\bibitem{Rusu_2010}
R.~B. Rusu, G.~Bradski, R.~Thibaux, , and J.~Hsu.
\newblock {Fast 3D recognition and pose using the viewpoint feature histogram}.
\newblock In {\em IROS}, 2010.

\bibitem{Sadanand_cvpr_2012}
S.~Sadanand and J.~Corso.
\newblock {Action bank: a high-level representation of activity in video}.
\newblock In {\em CVPR}, 2012.

\bibitem{Savva_2014_tog}
M.~Savva, A.~X. Chang, P.~Hanrahan, M.~Fisher, and M.~Niesner.
\newblock {SceneGrok: inferring action maps in 3D environments}.
\newblock {\em ACM Transactions on Graphics}, 2014.

\bibitem{CADstate3}
J.~Shan and S.~Akella.
\newblock {3D human action segmentation and recognition using pose kinetic
  energy}.
\newblock In {\em IEEE Workshop on Advanced Robotics and its Social Impacts},
  2014.

\bibitem{NIPS2014_twostream}
K.~Simonyan and A.~Zisserman.
\newblock {Two-stream convolutional networks for action recognition in videos}.
\newblock In {\em NIPS}, 2014.

\bibitem{Sun_cvpr_2012}
M.~Sun, P.~Kohli, and J.~Shotton.
\newblock {Conditional regression forests for human pose estimation}.
\newblock In {\em CVPR}, 2012.

\bibitem{sung_rgbdactivity_2012}
J.~Sung, C.~Ponce, B.~Selman, and A.~Saxena.
\newblock {Unstructured human activity detection from RGBD images}.
\newblock In {\em ICRA}, 2012.

\bibitem{hrf_iccv2013}
D.~Tang, T.-H. Yu, and T.-K. Kim.
\newblock {Real-time articulated hand pose estimation using semi-supervised
  transductive regression forests}.
\newblock In {\em ICCV}, 2013.

\bibitem{Ukita_2013}
N.~Ukita.
\newblock {Iterative action and pose recognition using global-and-pose features
  and action-specific models}.
\newblock In {\em ICCV Workshop}, 2013.

\bibitem{Vapnik2009544}
V.~Vapnik and A.~Vashist.
\newblock {A new learning paradigm: learning using privileged information}.
\newblock {\em Neural Networks}, 2009.

\bibitem{Vemuri_1986}
B.~C. Vemuri, A.~Mitiche, and J.~K. Aggarwal.
\newblock {Curvature-based representation of objects from range data}.
\newblock {\em Image and Vision Computing}, 1986.

\bibitem{Vieira_2012}
A.~W. Vieira, E.~R. Nascimento, G.~L. Oliveira, Z.~Liu, and M.~F. Campos.
\newblock {STOP: space-time occupancy patterns for 3D action recognition from
  depth map sequences}.
\newblock In {\em Progress in Pattern Recognition, Image Analysis, Computer
  Vision, and Applications}, 2012.

\bibitem{Chunyu_cvpr_2013}
C.~Wang, Y.~Wang, and A.~Yuille.
\newblock An approach to pose-based action recognition.
\newblock In {\em CVPR}, 2013.

\bibitem{ECCV2010_layout}
H.~Wang, S.~Gould, and D.~Koller.
\newblock {Discriminative learning with latent variables for cluttered indoor
  scene understanding}.
\newblock In {\em ECCV}, 2010.

\bibitem{Heng_2011_cvpr}
H.~Wang, A.~Klaser, C.~Schmid, and C.-L. Liu.
\newblock {Action recognition by dense trajectories}.
\newblock In {\em CVPR}, 2011.

\bibitem{Jiang_eccv_2012}
J.~Wang, Z.~Liu, J.~Chorowski, Z.~Chen, and Y.~Wu.
\newblock {Robust 3D action recognition with random occupancy patterns}.
\newblock In {\em ECCV}, 2012.

\bibitem{Jiang_tpami_2014}
J.~Wang, Z.~Liu, Y.~Wu, and J.~Yuan.
\newblock {Learning actionlet ensemble for 3D human action recognition}.
\newblock {\em TPAMI}, 2014.

\bibitem{wang2015cnn}
P.~Wang, W.~Li, Z.~Gao, J.~Zhang, C.~Tang, and P.~Ogunbona.
\newblock {Action recognition from depth maps using deep convolutional neural
  networks}.
\newblock {\em IEEE Transactions on Human Machine Systems}, 2015.

\bibitem{Wei_2013_ICCV}
P.~Wei, Y.~Zhao, N.~Zheng, and S.-C. Zhu.
\newblock {Modeling 4D human-object interactions for event and object
  recognition}.
\newblock In {\em ICCV}, 2013.

\bibitem{Willems_2008_eccv}
G.~Willems, T.~Tuytelaars, and L.~V. Gool.
\newblock {An efficient dense and scale-invariant spatio-temporal interest
  point detector}.
\newblock In {\em ECCV}, 2008.

\bibitem{Shandong_iccv_2011}
S.~Wu, O.~Oreifej, and M.~Shah.
\newblock {Action Recognition in videos acquired by a moving camera using
  Motion Decomposition of Lagrangian Particle Trajectories}.
\newblock In {\em ICCV}, 2011.

\bibitem{Lu_cvpr_2013}
L.~Xia and J.~K. Aggarwal.
\newblock {Spatio-temporal depth cuboid similarity feature for activity
  recognition using depth camera}.
\newblock In {\em CVPR}, 2013.

\bibitem{Xia_cvor_2012}
L.~Xia, C.-C. Chen, and J.~K. Aggarwal.
\newblock {View invariant human action recognition using histograms of 3D
  joints.}
\newblock In {\em CVPR Workshop}, 2012.

\bibitem{Yang_tcsvt_2015}
H.~Yang and I.~Patras.
\newblock {Privileged information-based conditional structured output
  regression forest for facial point detection}.
\newblock {\em TCSVT}, 2015.

\bibitem{PSSR2}
W.~Yang, Y.~Wang, and G.~Mori.
\newblock {Recognizing human actions from still images with latent poses}.
\newblock In {\em CVPR}, 2010.

\bibitem{Xiaodong_2014_cvpr}
X.~Yang and Y.~Tian.
\newblock {Super normal vector for activity recognition using depth sequences}.
\newblock In {\em CVPR}, 2014.

\bibitem{Yang_2012_acm}
X.~Yang, C.~Zhang, and Y.~Tian.
\newblock {Recognizing actions using depth motion maps-based histograms of
  oriented gradients}.
\newblock In {\em ACM International Conference on Multimedia}, 2012.

\bibitem{Yao_2011}
A.~Yao, J.~Gall, G.~Fanelli, and L.~V. Gool.
\newblock {Does human action recognition benefit from pose estimation?}
\newblock In {\em BMVC}, 2011.

\bibitem{Yu_2013_cvpr}
T.-H. Yu, T.-K. Kim, and R.~Cipolla.
\newblock {Unconstrained monocular 3D human pose estimation by action detection
  and cross-modality regression Forest}.
\newblock In {\em CVPR}, 2013.

\bibitem{Zanfir_2013_ICCV}
M.~Zanfir, M.~Leordeanu, and C.~Sminchisescu.
\newblock {The moving pose: an efficient 3D kinematics descriptor for
  low-latency action recognition and detection}.
\newblock In {\em ICCV}, 2013.

\bibitem{Zhang_2011}
H.~Zhang and L.~E. Parker.
\newblock {4-dimensional local spatio-temporal features for human activity
  recognition}.
\newblock In {\em IROS}, 2011.

\bibitem{Yang_cvpr_2015}
Y.~Zhou, B.~Ni, R.~Hong, M.~Wang, and Q.~Tian.
\newblock {Interaction part mining: a mid-level approach for fine-grained
  action recognition}.
\newblock In {\em CVPR}, 2015.

\bibitem{CVPR2013_daction2}
Y.~Zhu, W.~Chen, and G.~Guo.
\newblock {Fusing spatiotemporal features and joints for 3D action
  recognition}.
\newblock In {\em CVPR Workshop}, 2013.

\bibitem{Zhu_2014_IVC}
Y.~Zhu, W.~Chen, and G.~Guo.
\newblock {Evaluating spatiotemporal interest point features for depth-based
  action recognition}.
\newblock {\em Image and Vision Computing}, 2014.

\end{thebibliography}
}

\end{document}